# Multiobjective Optimization of Classifiers by Means of 3-D Convex Hull Based Evolutionary Algorithms

Jiaqi Zhao, Vitor Basto Fernandes, Licheng Jiao, Iryna Yevseyeva, Asep Maulana, Rui Li, Thomas Bäck, and Michael T. M. Emmerich

**Abstract**

Finding a good classifier is a multiobjective optimization problem with different error rates and the costs to be minimized. The receiver operating characteristic is widely used in the machine learning community to analyze the performance of parametric classifiers or sets of Pareto optimal classifiers. In order to directly compare two sets of classifiers the area (or volume) under the convex hull can be used as a scalar indicator for the performance of a set of classifiers in receiver operating characteristic space.

Recently, the convex hull based multiobjective genetic programming algorithm was proposed and successfully applied to maximize the convex hull area for binary classification problems. The contribution of this paper is to extend this algorithm for dealing with higher dimensional problem formulations. In particular, we discuss problems where parsimony (or classifier complexity) is stated as a third objective and multi-class classification with three different true classification rates to be maximized.

The design of the algorithm proposed in this paper is inspired by indicator-based evolutionary algorithms, where first a performance indicator for a solution set is established and then a selection operator is designed that complies with the performance indicator. In this case, the performance indicator will be the volume under the convex hull. The algorithm is tested and analyzed in a proof of concept study on different benchmarks that are designed for measuring its capability to capture relevant parts of a convex hull.

Further benchmark and application studies on email classification and feature selection round up the analysis and assess robustness and usefulness of the new algorithm in real world settings.

## I. INTRODUCTION

Classification is one of the most common problems in machine learning. The task of classification is to assign instances in a data set to target classes based on classifiers previously learned. The receiver operating characteristic (ROC) graph is a technique for visualizing, organizing and selecting classifiers based on their performance [1]. ROC graphs are usually used to evaluate and compare the performance of classifiers and they also have properties that make them especially useful for domains with skewed class distributions and different classes of problems that assign costs to misclassification. Originating from the field of object classification in radar images, ROC analysis has become increasingly important in many other areas with cost sensitive classification and/or unbalanced data distribution, such as medical decision making [2], signal detection [3], diagnostic systems [4].

Jiaqi Zhao and Licheng Jiao are with the Key Laboratory of Intelligent Perception and Image Understanding of the Ministry of Education, International Research Center for Intelligent Perception and Computation, Xidian University, Xi'an Shaanxi Province 710071, China (e-mails:jiaqizhao88@126.com, lchjiao@mail.xidian.edu.cn).

Vitor Basto Fernandes is with the School of Technology and Management, Computer Science and Communications Research Centre, Polytechnic Institute of Leiria, 2411-901 Leiria, Portugal. (e-mail: vitor.fernandes@ipleiria.pt).

Asep Maulana, Thomas Bäck and Michael T.M. Emmerich are with the Leiden Institute for Advanced Computer Science (LIACS), Leiden University, 2333 CA Leiden, Netherlands. (e-mails: {a.asep.maulana, t.h.w.baeck, m.t.m.emmerich}@liacs.leidenuniv.nl).

Rui Li is with Microsoft Research Asia, Beijing 100190, China (e-mail: lerain@gmail.com).

Iryna Yevseyeva is with Cyber Security Research Institute, School of Computing Science, Newcastle University, NE1 7RU, Newcastle-upon-Tyne, UK (e-mail:iryna.yevseyeva@newcastle.ac.uk).

More recently, research has drawn attention to ROC convex hull (ROCCH) analysis that covers potentially optimal points for a given set of classifiers [1]. ROCCH makes use of the finding that two hard classifiers can be combined to a classifier that has characteristics in ROC space that correspond to linear combinations of the characteristics of single classifiers and thus these linear combinations do not have to be explicitly represented in ROC space. A performance indicator for sets of hard binary classifiers that is compliant with the improvement of ROCCH is the area under the convex hull (AUC).

Some evolutionary multiobjective optimization algorithms (EMOAs) [5] have been applied to machine learning and image processing areas [6]. The maximization of the performance of ROC representations with respect to this indicator has been subject to a recent study by Wang et al. [7], who showed that the proposed algorithm, CH-EMOA, is capable to show a strong performance for improving ROCCH with respect to AUC as compared to using state-of-the-art EMOAs (NSGA-II [36], SPEA2 [40], MOEA/D [41], and SMS-EMOA [38]) for the same task.

So far algorithms that seek to maximize ROCCH performance have only focused on the problem of optimizing binary classifiers with respect to two criteria. There is however an increasing interest in extending ROCCH performance analysis to more than two criteria. In this research we consider the following additional objectives:

- Complexity minimization: The objective here is to find models with maximum simplicity (parsimony) or minimum computational costs. It can be described as the number of rules defining a classifier in proportion of the number of possible rules. In the context of feature selection this objective favors classifiers with a small number of features.
- Three-class classification: There can be three or more classes and for each class the error rates need to be minimized. A simplified ROC is estimated from a three-class classifier by only considering the rate of every correctly classified category [8], which ensures that good classifiers tend to result in better ROCCH than poorer ones.

The contribution of this article is to extend the CH-EMOA to deal with 3-D ROCCH problems, by considering these additional objectives for the classifier. As a new algorithm, the 3-D convex hull evolutionary multiobjective algorithm (3DCH-EMOA) can deal with many additional problems in the machine learning field. Moreover, we analyze and assess the performance of the algorithm in different studies on, partly new, academic problems and practical applications. To analyze the capability of different algorithms to maximize convex hull volume, in a more fundamental study two sets of test problems named ZEJD and ZED are proposed and the capability of 3DCH-EMOA to capture only the convex part of a Pareto front is assessed. These benchmarks are accompanied by standard benchmarks from the machine learning community. Besides, we include a study on SPAM filter design and a study on feature selection. While in the SPAM filter study the number of rules determines the complexity objective in terms of number of used rules, the goal in feature selection is to reduce the number of features. 3DCH-EMOA is compared with state-of-the-art EMOAs such as NSGA-II [36], GDE3 [39], SPEA2 [40], MOEA/D [41], and SMS-EMOA [38]on the test problems.

This paper is organized as follows: the related work is outlined in Section II and the background of 3-D ROC surfaces and the theory of multiobjective optimization are introduced in Section III. We describe the framework of the 3DCH-EMOA algorithm in Section IV and experimental results on ZJED and ZED benchmarks test problems are described and discussed in Section V. The description of the SPAM filter application and experimental results are shown in Section VI. In Section VII the experimental results of feature selection for binary and three-class classification are discussed. Section VIII provides the conclusions and a discussion on the important aspects and future perspectives of this work.

## II. Related work

### A. ROCCH in classification

The receiver operating characteristic convex hull (ROCCH) covers all the potential optimal classifiers in a given set of classifiers. Therefore, the aim of ROCCH maximization is to find a group of classifiers that perform well as a set. Despite the fact that ROCCH is an important topic in classification, there is not much work focusing on how to maximize the ROCCH. A reason for this could be that this is a relatively complex task compared to approaches that assess performance of a classifier by means of a single number. However, the additional gain in information about the trade-off between different objectives and the possibilities it offers for online adjustments of classifiers should justify the development of more mature methods for ROCCH maximization. The set of existing methods could be partitioned into two categories: one is ROC geometric analysis based machine learning methods and the other one is evolutionary multiobjective optimization methods.

ROCCH maximization problems were first described in [9]. One approach is to identify portions of the ROCCH to use iso-performance lines [1] that are translated from operating conditions of classifiers. Suitable classifiers for data sets with different skewed class distribution or misclassification costs can be chosen based on these iso-performance lines. In addition, a rule learning mechanism is described in [10] and in [11]. It combines rule sets to produce instance scores indicating the likelihood that an instance belongs to a given class, which induces decision rules in ROC space. In the above methods a straightforward way was used to analyze the geometrical properties of ROC curves to generate decision rules. However, the procedure is not effective and easily gets trapped in local optima.

In [12], a method for detecting and repairing concavities in ROC curves is studied. In that work, a point in the concavity is mirrored to a better point. This way the original ROC curve can be transformed to a ROC curve that performs better. The Neyman-Pearson lemma is introduced to the context of ROCCH in [13], which is the theoretical basis for finding the optimal combination of classifiers to maximize the ROCCH. This method not only focuses on repairing the concavity but it also improves the ROC curve, which is different with [12]. For a given set of rules, the method can combine the rules using *and* and *or* to get the optimum rule subset efficiently. But it can not generate new rules in the global rule set. More recently, ROCCER was proposed in [11]. It is reported to be less dependent on the previously induced rules.

Recently also multiobjective optimization techniques to maximize ROCCH received attention. The ROCCH maximization problem is a special case of a multiobjective optimization problem, because the minimization of false positive rates and maximization of true positive rates can be viewed as conflicting objectives, and the parameters of a classifier can be viewed as decision variables. In [14], non-dominated decision trees were developed, which are used to support the decision which classifier to choose. A multiobjective genetic programming approach was proposed to envelop the alternative Pareto optimal decision trees. However, it is not a general method for ROCCH maximization because it only pays attention to cost sensitive classification problem.

The Pareto front of multiobjective genetic programming is used to maximize the accuracy of each minority class with unbalanced data set in [15]. Moreover, in [16], the technique of multiobjective optimization genetic programming is employed to evolve diverse ensembles to maximize the classification performance for unbalanced data.

Some more evolutionary multiobjective optimization algorithms (EMOAs) have been combined with genetic programming to maximize ROC performance in [17]. Although they have been used in ROCCH maximization, these techniques do not consider special characteristics of ROCCH. This is done in convex hull multiobjective genetic programming (CH-MOGP), which is proposed in [7]. CH-MOGP is a multiobjective indicator-based evolutionary algorithm (IBEA) using the area under the convex hull (AUC) as a performance indicator for guiding the search. It has been compared to other state-of-the-art methods and showed the best performance for binary classifiers on the UCI benchmark [18]. However, it is so far limited to binary classifiers and does not take into account additional objectives, such as parsimony. The new research discussed in the following takes the algorithm design and analysis to the next level by extending the concepts introduced in CH-MOGP to problems with three objective functions, and by providing a detailed analysis for the performance of the resulting algorithm 3DCH-EMOA.

*B. ROC from multiclass classifiers*

The ROC graph is often used to evaluate and optimize the performance of a binary classifier. The area under the ROC convex hull (AUC) has become a standard performance evaluation criterion in binary pattern recognition problems and it has been widely used to compare different classifiers independently of priors of distribution and costs of misclassification [8]. However, the AUC measure is only applicable to the binary classification problem.

The ROC curve is extended to multi-class classification problems in [19]. Instead of two-dimensional points, now optimal classifiers are represented by points in an $n$-dimensional ROC space. A more recent example is the ROC hyper-surface (see, e.g., [20]) that comprises the non-dominated boundary of the convex hull in $n$-dimensional ROC space. Many desirable properties, such as the possibility to repair concavities, are preserved by the ROC hyper-surface.

It has been shown that a multi-class classifier with good ROC hyper-surface can lead to classifiers suitable for various class distributions and misclassification costs via re-optimization of its output matrix [20]. However, due to the increase of the dimensionality of the ROC space, achieving the optimal ROC hyper-surface is even more difficult than achieving the optimal ROC curve [21]. Hence, AUC has to be extended to multi-class classifiers to enable the comparison between classifiers on the basis of more detailed information about the performance characteristics. A straightforward way to generalize AUC is the volume under the ROC (a surface in 3-D and a hyper-surface in $n$-D) surface (VUS) [22]. Methods from computational geometry can be used to efficiently compute the VUS, but, as compared to the 2-D case, this is relatively complicated [23]. Moreover, the VUS value of a "near guessing" classifier is almost the same as a "near perfect" classifier for more than two classes (see [24]). However, alternative definitions of VUS have been proposed where this problem does not occur [8]. This will be discussed later in more detail.

The VUS from multi-class classifiers is simplified by considering the AUC between each class and all of the other classes (one vs. all strategy) in [25], turning the problem to a problem that considers a set of binary classifiers. This approach ignores higher order interactions and depends only on class distribution priors and misclassification costs. In [26], multi-class area under convex hull (MAUC) has been proposed as a simpler generalization of AUC for multi-class classification problems. MAUC has been widely used in recent work [21], [27]. MAUC is estimated by averaging the AUC between all pairs of classes. This leads to a quadratically growing number of required comparisons in the number of classes [8].

The VUS of three-class classifiers has been studied in [28] and [29]. The simplified ROC is estimated from a multiclass classifier by only considering the rate of every correctly classified category [8], which ensures that good classifiers tend to result in better VUS than bad classifiers. The theory of ROCCH can easily be applied to multi-class classifiers in this way and many results, such as the possibility to heal concavities, can be generalized. In this paper, we therefore focus on finding (sets of) classifiers with optimal VUS considering the simplified ROC proposed in [8].

*C. Convex Hull based EMOA*

In the past, convex hull based selection operators were employed in EMOA to maintain a well-distributed set or make the non-dominated sorting more effective (cf. [30], [31]).

In [32] a multiobjective evolutionary algorithm based on the properties of the convex hulls defined in the 2-D ROC space was proposed, which was applied to determine a set of fuzzy rule-based binary classifiers with different trade-offs between false positive rate (fpr) and true positive rate (tpr). NSGA-II was used to generate an approximation of a Pareto front composed of genetic fuzzy classifiers with different trade-offs among sensitivity, specificity, and interpretability in [33]. After projecting the overall Pareto front onto the ROC space, ROC convex hull method was used to determine the potential optimal classifiers on the ROC plane.

In a recent research article Wang et al. discuss the optimization of binary classifiers based on receiver operator characteristic [17]. They use an indicator-based selection operator that seeks to maximize the AUC. This strategy proves to be very effective for finding relevant parts of the Pareto front. Here, we therefore adopt the strategy, but lift it to the next level by considering multidimensional objective spaces.

## III. 3-D ROC AND MULTIOBJECTIVE OPTIMIZATION

Our study aims at looking at different types of three-objective problems, including (1) maximization of true category classification in three-class classification problems and (2) binary classification problems with parsimony or complexity (number of rules, number of features) as a third objective. For this we propose a multiobjective optimization algorithm to maximize ROCCH in 3-D.

|  | | True class | |
|---|---|---|---|
|  | | P | N |
| Predicted class | P | True Positives | False Positives |
|  | N | False Negatives | True Negatives |

Fig. 1.  Confusion matrix of binary classifiers

The ROC curve of a binary classifier is defined by a two-by-two confusion matrix which describes the relationship between the true labels and predicted labels from a classifier. An example of a confusion matrix is shown in Fig. 1. There are four possible outcomes with binary classifiers in a confusion matrix. It is a true positive (TP), if a positive instance is classified as positive. We call it false negative (FN), if a positive instance is classified as negative. If a negative instance is correctly classified we call it a true negative (TN), else we call it a false positive (FP).

Let fpr = FP/(TN+FP) be the false positive rate and fnr = FN/(TP+FN) be the false negative rate. Since no perfect classifier exists, and fpr and fnr are conflicting with each other, ROC graphs are used to depict the trade-off between them. ROC graphs are two-dimensional graphs in which the fpr is plotted on the X axis and fnr is plotted on the Y axis.

Basically, finding a set of optimal classifiers is a bi-objective problem. Next we will look at related problem definitions with more than two objective functions: Firstly, classifier complexity (or simply 'complexity') is discussed as a third objective function and secondly the three-class classification problem is introduced.

### A. Complexity as a third objective function

Besides fpr and fnr, we define the third objective as complexity of the classifier. In classifiers with feature selection the number of features or used features rate (ufr which is defined in Eq. 1) can be used as the third objective, whereas in rule-based classifiers the number of rules from a rule base or used rules rate (urr which is defined in Eq. 2) can be used as a complexity objective. In the general case we will use the term *classifier complexity ratio (ccr)* for both of them.

$$ufr = \frac{number\ of\ used\ features}{total\ number\ of\ features} \tag{1}$$

$$urr = \frac{number\ of\ used\ rules}{total\ number\ of\ rules} \tag{2}$$

The ccr will obtain its maximum in the value 1.0 (all features or, respectively, rules are used) and its minimum in the value of 0.0 (no feature or, respectively, rule is used). The computational cost of a classifier with a high ccr is considered to be higher than that of a lower ccr, while the classifier with the lower ccr should be preferred given the false positive and false negative rates are the same. It is always possible to construct a classifier the characteristic of which is a convex combination of the original classifier by means of randomization, and in some cases, by means of finding classifiers that combine features, or respectively, rules of two classifiers. We name the ROC space with complexity of binary classifiers as a third axis in the augmented ROC space.

In augmented ROC space for complexity classifiers, fpr is plotted on X axis, fnr is plotted on Y axis, and ccr is plotted on the Z axis, which is depicted in Fig. 2. Normally, ccr, fpr and fnr are conflicting with each other. A low value of ccr means low complexity of the classifier, i.e. there are less rules for rule-based classifier or less features for feature selection problem, which would result in a poor performance of fpr and fnr. The new proposed algorithm aims at finding trade-offs among the three objectives.

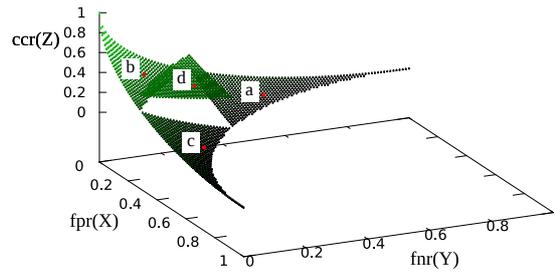

Fig. 2.  An example of an augmented ROC graph with complexity of binary classifiers as a third axis

Several points in augmented ROC space are important to note. The point $(0,0,0)$ represents the strategy of never issuing a wrong classification and the classifier with a cost of zero. This point represents a perfect classifier and a classifier corresponding to such a point does not exist in reality but can be approximated as closely as possible. The point $(1,0,0)$ represents the strategy of issuing all the instances as negative by a classifier whose complexity is zero. The point $(0,1,0)$ represents a classifier that predicts all instances as positive without using any rules or features. In a similar way, predicting all of the instances as negative with all the rules or features results in the point $(1,0,1)$. The point $(0,1,1)$ can be obtained by predicting all of the instances as positive with all the rules or features. For all points in $\{(1,0,0),\ (0,1,0),\ (1,0,1),\ (0,1,1)\ \}$ a classifier can be constructed. The surface of fpr+fnr=1 represents randomly guessing classifiers, which is shown in Fig. 3. The classifiers which we search for should be in the space of fpr+fnr<1. In this paper, we treat the points $(1,0,0)$, $(0,1,0)$, $(1,0,1)$ and $(0,1,1)$ as reference points to build ROCCH and to calculate the VUS in augmented ROC space. For a set of randomly guessing classifiers the VUS will be $0$, and for a set of perfect classifiers the VUS will be $0.5$ in augmented ROC space.

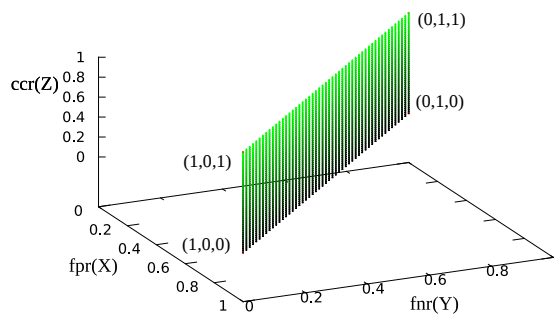

Fig. 3.  An example of an augmented ROC graph for random binary classifiers

Every complexity augmented binary classifier can be mapped to the augmented ROC space. The ROCCH is the collection of all potentially optimal classifiers in a given set of classifiers. Furthermore, a classifier is potentially optimal if and only if it lies on the augmented ROCCH of the set of points in ROC space for complexity classifiers. In Fig. 2 the points $a, b, c$ are on the ROC surface and the point $d$ is above it. $a, b, c$ represent potentially optimal classifiers and $d$ is not. Any virtual classifier can be constructed by combining two or more classifiers on the ROC surface. The newly proposed method aims at maximizing ROCCH by maximizing VUS.

### B. ROCCH for three-class classification

Basically, ROC analysis concerns the confusion matrix for the outputs of a classifier, which is defined by a three-by-three matrix for a three-class classifier as it is shown in Fig. 4. In order to explain it clearly, we define the three classes as class A, B and C. There are nine possible outcomes with a three-class classification problem. An outcome is called a True A (TA) if an instance A is classified as A, we call it a False Ab (FAb) if an instance B is classified as A and we call it a False Ac (FAc) if an instance C is classified as A. In a similar way, we define a False Ba (FBa), a True B (TB), a False Bc (FBc), a False Ca (FCa) and a False Cb (FCb) and True C (TC) which are described in Fig. 4. Six variables are required to describe the confusion matrix exactly. The problem is simplified by only considering the diagonal elements of the matrix in [8]. Let tar = TA/(TA+FBa+FCa) be the true A rate, tbr = TB/(TB+FAb+FCb) be the true B rate and tcr = TC/(TC+FAc+FBc) be the true C rate. In this way, the problem of three-class classification turns out to maximize the tar, tbr and tcr simultaneously. However, because they are normally conflicting each other, the newly proposed method aims at finding trade-offs among them.

|  | True class | | |
|---|---|---|---|
|  | A | B | C |
| A | True A | False Ab | False Ac |
| Predicted class B | False Ba | True B | False Bc |
| C | False Ca | False Cb | True C |

Fig. 4. Confusion matrix of three-class classifiers

ROC graphs for three-class classifiers are three-dimensional graphs in which tar is plotted on the X axis, tbr is plotted on the Y axis and tcr is plotted on the Z axis, as it is shown in Fig. 5.

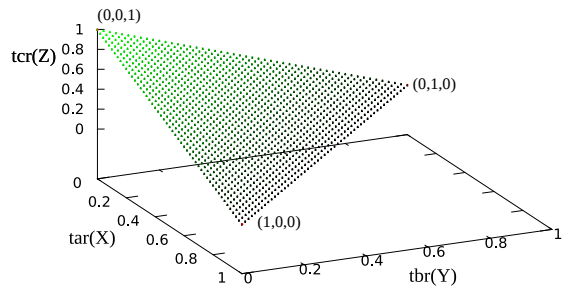

Fig. 5. Random classification performance of the three-class ROC

Several points in three-class ROC space are important to note. The lower point (0,0,0) represents the strategy of never issuing a right classification, all of the outcomes of such a classifier are wrong labels. The opposite strategy, the upper point (1,1,1) represents the strategy of never issuing a wrong classification, which represents a perfect classifier. Both of these classifiers can normally not be constructed. The point (1,0,0) represents the case where all instances of the true class A are classified as A but no true instances of class B and C are correctly classified. The point (1,0,0) can be achieved by classifying all the instances as class A. The point (0,1,0) represents the case where all instances of the true class B are classified as B but no true instances of class A and C are correctly classified. And the point (0,0,1) represents the case where all instances of the true class C are classified as C but no true class A and B are correctly classified. The points (1,0,0), (0,1,0) and (0,0,1) are fixed points on three-class ROC graphs. We treat them as reference points to build ROCCH and to calculate the VUS under 3-D ROC space of three-class classification case. The value of VUS will be zero for random guessing classifiers and the value of a perfect classifier is $5/6$. The value of VUS of common classifiers is in the range $[0, 5/6]$, a large value always corresponding to classifiers with good performance.

The surface tar+tbr+tcr=1 on the 3-D ROC space represents the strategy of randomly guessing a class for the instance, see e.g. Fig. 5. For example, if a classifier randomly guesses the A class 0.4 times on average, the B class 0.4 times on average and the C class 0.2 of the total time used for classification, 40 percent of the class A instances can be correctly classified, 40 percent of class B instances will be correctly classified and only 20 percent of class C instances will be correctly classified. This yields the point (0.4,0.4,0.2) in 3-D ROC space. If it guesses the instances as class A all the times on average and no times for class B and class C, 100 percent of class A instances will be correctly classified and 0 percent of both class B and class C will be correctly classified. This yields the point (1,0,0) in 3-D ROC space. Any classifiers on the triangle surface in 3-D ROC space may be said to have no information about the class. A classifier that appears in the lower triangle surface performs worse than random guessing. A more effective classifier produces the ROC surface above the triangle surface.

In the following, we will give an example of using 3-D ROC to search optimal classifiers for data sets with different class distributions. One important goal of classification problems is to maximize the total accuracy. Suppose the priori probability of the data distribution of every class is denoted as p(a), p(b) and p(c), where p(a)+p(b)+p(c)=1. The accuracy (acc) of a classifier can then be represented as in Eq. 3.

$$acc = p(a) \cdot tar + p(b) \cdot tbr + p(c) \cdot tcr \tag{3}$$

While considering cost-sensitive problems, the cost function is simplified by only considering the cost of misclassified instances for each class. We denote the cost of each class as c(a), c(b) and c(c). Here c(a) represents the cost of classifying an instance of class $a$ as class $b$ or class $c$, c(b) and c(c) are defined in a similar way. The total cost is described in Eq. 4, in which $N$ is the total number of instances.

$$
\begin{aligned}
cost = {} & N \cdot p(a) \cdot c(a) \cdot (1 - tar) \\
& + N \cdot p(b) \cdot c(b) \cdot (1 - tbr) \\
& + N \cdot p(c) \cdot c(c) \cdot (1 - tcr)
\end{aligned}
\tag{4}
$$

If there are two points (tar1,tbr1,tcr1) and (tar2,tbr2,tcr2) that have the same accuracy or cost, we can get an extended iso-performance surface as described in Eq. 5 and Eq. 6. Both Eq. 5 and Eq. 6 are surface functions which are named as iso-performance surface in ROC analysis theory [1].

$$
\begin{aligned}
& p(a)(tar1 - tar2) \\
& + p(b)(tbr1 - tbr2) \\
& + p(c)(tcr1 - tcr2) = 0
\end{aligned}
\tag{5}
$$

$$
\begin{aligned}
& p(a) \cdot c(a)(tar1 - tar2) \\
& + p(b) \cdot c(b)(tbr1 - tbr2) \\
& + p(c) \cdot c(c)(tcr1 - tcr2) = 0
\end{aligned}
\tag{6}
$$

All the classifiers corresponding to the points on an iso-performance surface have the same expected accuracy or cost. Moreover, each set of class distribution defines a family of iso-performance surfaces. By moving the iso-performance surface until it gets in contact with a point in ROCCH, the joint point represents a classifier which can produce highest accuracy or lowest cost under that condition. Generally, the larger value of VUS the better performance the ROCCH will be.

Sometimes the classifier with desired performance is not in the set of available classifiers, but lies between two of them. The classifier with desired performance can be obtained by sampling the decisions of each classifier. The sampling ratio is determined by the position where the resulting classifier lies. The way to find a desired optimal classifier is similar to ROCCH for binary classifiers which is described in [1].

Imprecise distribution information of data defines a range of parameters for iso-performance lines (surfaces) and the range of lines (surfaces) will intersect a segment of 3D ROCCH. If the segment defined by a range of lines corresponds to a single point in ROC space, then there is no sensitivity to the distribution assumptions, otherwise the ROCCH is sensitive to the distribution assumptions. In order to improve the robustness of ROCCH not only the VUS should be maximized but also its structure should be optimized. Usually, the more uniform the distribution of points in the ROCCH, the more nonsensitive the ROCCH is. The gini coefficient [34] is used to evaluate the uniformity of the solution of the test functions in this paper, and the details will be discussed later.

### C. ROCCH maximization and multiobjective optimization

The goals of augmented ROCCH and three-class ROCCH maximization are to find a group of classifiers that approximate the perfect point (0,0,0) for complexity binary classifiers and (1,1,1) for three-class classifiers, respectively. The ROCCH maximization problems turn out to be minimization multiobjective optimization problems by multiplying each objective function with minus one as described in Eq. 7.

$$\min F(x) = (f_1(x), f_2(x), f_3(x))$$
$$\text{subject to} \quad x \in \Omega \tag{7}$$

In Eq. 7, $x$ is a set of classifiers, $\Omega$ is the solution space, i.e. the set of all possible classifiers, and $F(x)$ is a vector function for ROC distribution. In the theory of multiobjective optimization, dominance is an important concept which is defined as: Let $\mu = (\mu_1, \mu_2, \mu_3)$, $\nu = (\nu_1, \nu_2, \nu_3)$ be two vectors, $\nu$ is said to dominate $\mu$ if $\nu_i \leq \mu_i$ for all $i = 1, 2, 3$, and $\nu \neq \mu$, this is denoted as $\nu \prec \mu$. $\nu$ and $\mu$ are incomparable if $\nu$ and $\mu$ do not dominate each other [7]. The Pareto set (PS) is the collection of all the Pareto optimal points in decision space, i.e. of all points $x \in \Omega$ with no $x' \in \Omega$ such that $f(x') \prec f(x)$. The Pareto front is the set of all objective vectors of points in PS in objective space $PF = \{F(x) \mid x \in PS\}$ [17].

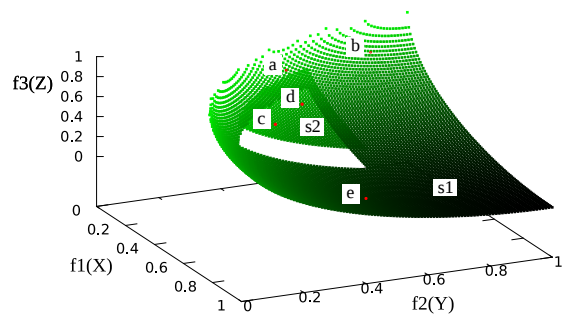

Fig. 6. Convex hull and Pareto front in 3-D ROC space

A special approach based on ROCCH is proposed in [7] to solve the ROC maximization problem for binary classification. The convex hull is different from the Pareto front though they are similar to each other. In the example of Fig. 6 points $a, b, c, d, e$ are non-dominated points in traditional multiobjective optimization algorithms. However, only points $a, b$ and $e$ are on the convex hull. This is the special character of the ROCCH maximization problem and new strategies need to be researched to deal with it. We call these problems ROCCH optimization problems. While in [7] binary classifiers were considered, here we consider complexity augmented binary classifiers and three-class classifiers and in the context of 3-D ROCCH multiobjective maximization.

*D. The motivation and ideas for 3-D ROCCH maximization*

There are two key steps of any EMOAs, one is how to rank the population and the other one is selecting solutions that survive to the next generation. The ranking approach of NSGA-II, probably the most popular EMOA used up-to-date, includes two steps. First sorting the population into several levels indicating the priority, and second designing a selection scheme which is used to choose winner individuals to survive from populations at the same level. While dealing with the 3-D ROCCH maximization problem, 3-D convex hull based sorting without redundancy is used to sort the population, which makes the diversity increase fast in the evolutionary process. A volume contribution selection scheme is adopted to calculate the contribution to the volume of 3-D ROCCH of each solution. Based on our insight, the VUS based selection strategy has a better performance than hyper-volume or crowding-distance to deal with the 3-D ROCCH maximization problem, because the special characteristic of ROCCH problem has been taken into consideration. The excellent performance will be shown in the experimental results.

## IV. 3-D CONVEX HULL BASED EVOLUTIONARY MULTIOBJECTIVE OPTIMIZATION

In this section, the newly proposed 3-D convex hull based evolutionary multiobjective optimization algorithm (3DCH-EMOA) is described. There are several differences between 3DCH-EMOA and other EMOAs. Firstly, 3-D convex hull (3DCH) based sorting without redundancy approach is used to rank the individuals into several priority levels as proposed in [7]. Secondly, a new strategy of VUS contribution is designed to rank the solutions in the same priority level. Thirdly, a non-descending $(\mu + 1)$ selection strategy is employed in this paper, rather than the approximate $(\mu + \mu)$ scheme in CH-MOGP [7]. The 3-D convex hull is built by the quickhull algorithm which is proposed in [37]. The quickhull algorithm is used to sort the populations in different levels and calculate the VUS contribution for every point on the convex hull. The details of these strategies are discussed next.

*A. 3DCH based sorting without redundancy*

Convex hull based sorting without redundancy strategy was first proposed in [7], which has a good performance to deal with binary classification problems. The convex hull based sorting approach is extended to three dimensions in this paper. The strategy works effectively, not only because it can maintain the diversity of the population, but also because it takes into consideration the properties of ROCCH. With this strategy, if there are not enough non-redundant solutions to fill the whole population, the redundant solutions which are preserved in the archive should be randomly selected in population. It will be shown in Section V that the strategy can preserve the diversity of the population by keeping non-redundant solutions with bad performance and discarding the redundant solutions even with good performance. In addition, the non-redundancy strategy can avoid the solutions at the convex hull being copied many times at the selection step of the algorithm.

The framework of 3DCH based without redundancy sorting is described in Algorithm 1. At first the solution set $Q$ is divided into two parts, one is the redundancy solution set $Q_r$ the other is the non-redundancy solution set $Q_{nr}$. The redundancy solution set $Q_r$ will be assigned to the last level of priority of the solution set and the non-redundancy solution set $Q_{nr}$ will be assigned into different priority levels by a convex hull based ranking method. Before ranking the non-redundancy solution set $Q_{nr}$, a reference point set $R$ should be merged with it and a set of candidate points of convex hull $T$ is constructed. The 3-D quickhull algorithm [37] is adopted to build the convex hull with the candidate points set, which is widely used in 3-D convex hull related applications. The points (solutions) on the convex hull surface are considered as the current Pareto set and the first level of the Pareto front (note that all points can be constructed by means of a convex combination of classifiers in the set) and the remaining points will be used to build the new convex hull for the next level of Pareto front. Usually, there are several levels of solutions in the beginning of the algorithm and the number of levels will converge to one with the evolution of the population.

---

**Algorithm 1** 3DCH based sorting without redundancy $(Q, R)$

---

**Require:**    $Q \neq \varnothing$.
     $Q$ is a solution set.
     $Q_r$ is the redundancy solution set.
     $Q_{nr}$ is the non-redundancy solution set.
     $R$ is the set of reference points.
     $F$ is the ranked solution set in different levels.
     $F_0$ represents the Pareto front of the population.
**Ensure:**    3DCH based sorting without redundancy
  1: $Q = Q_r + Q_{nr}$
  2: $T = Q_{nr} \cup R$
  3: $F_0 = $ 3D quickhull algorithm(T) [37]
  4: $Q_{nr} = Q_{nr} - F_0$
  5: i=1
  6: **while** $Q_{nr} \neq \varnothing$ **do**
  7:     $T = Q_{nr} \cup R$
  8:     $F_i = $ 3D quickhull algorithm(T)
  9:     $Q_{nr} = Q_{nr} - F_i$
10:     $i = i + 1$
11: **end while**
12: $F_i = Q_r$
13: **return**   the ranked solution set F

---

An example of the result of 3DCH based sorting without redundancy is given in Fig. 6. The surfaces s1 and s2 represent two levels of solutions of different priority, the solutions on surface s1 are better than those on surface s2 and the solutions on surface s1 have much more opportunity to survive to the next generation than those on s2.

After ranking the individuals into different priority levels, another question arises such as how to analyze the importance of individuals in the same priority level. As the redundant solutions have no additional information about the population, the algorithm selects some of them to survive to the next generation randomly. If there are too many non-redundant solutions to fill the new population, the contribution to the VUS will be used as metric measure to rank the individuals in the same level, and only the high contribution individuals will survive. The details of the contribution of VUS will be described in the next section.

### B. VUS contribution selection scheme

In this section, we describe the VUS contribution indicator to evaluate the importance of individuals within the same priority level. We hypothesize that the new VUS contribution indicator is a more efficient strategy to maximize the volume under the 3-D convex hull when compared to the hyper-volume based contribution [38] or crowding distance indicator [36]. To calculate the contribution of an individual, a new convex hull should be built by subtracting from the total population volume the volume of the population without the individual, as shown in Eq. 8.

$$\Delta VUS_i = VUS(Q) - VUS(Q - \{q_i\})$$
$$i = 1, ..., m \qquad (8)$$

The procedure of calculating the VUS contribution for the non-redundancy solution set $Q_{nr}$ is given in Algorithm 2. After calculating the contribution to the VUS of each individual in $Q_{nr}$, the individuals in the same priority level can be ranked by the volume of $\Delta$VUS. The larger the volume of the contribution to VUS the more important the individual will be.

---

**Algorithm 2** $\Delta$VUS$(Q_{nr}, R)$

---

**Require:**   $Q_{nr} \neq \varnothing$
        $Q_{nr}$ is the non-redundancy solution set
        $R$ is a reference points set
**Ensure:**   $\Delta$VUS
  1: $m = sizeof(Q_{nr})$
  2: $P = Q_{nr} \cup R$
  3: $Volume_{all} = VUS(P)$ [37]
  4: **for all** i=1:m **do**
  5:    $q_i \leftarrow Q_{nr}(i)$
  6:    $\Delta VUS_i = Volume_{all} - VUS(P - \{q_i\})$
  7: **end for**
  8: **return**  $\Delta$VUS

---

### C. 3DCH-EMOA

The framework of 3DCH-EMOA is described in Algorithm 3, which is inspired by indicator-based evolutionary algorithms. To optimize the multiple objectives on the convex hull space the initial population $Q_0$ should be built randomly. The scheme of a $(\mu + 1)$ evolutionary algorithm is adopted within 3DCH-EMOA to maximize the ROC performance, rather than the $(\mu + \mu)$ scheme used in CH-MOGP [17], which causes the algorithm to converge too quickly. For each iteration there is only one offspring produced by the evolutionary operators and to keep the size of population constant, the least performing individual should be deleted. The non-descending reduce strategy given in Algorithm 4 is adopted in this method to remove an individual from the population, which ensures that the population becomes better with the evolution of the generation.

---

**Algorithm 3** 3DCH based EMOA $(Max, N)$

---

**Require:**   $Max > 0, N > 0$
        $Max$ is the maximum number of evaluations
        $N$ is the population size
**Ensure:**   3DCH-EMOA
  1: $Q_0 = initialization()$
  2: $t_0 = 0$
  3: $m = 0$
  4: **while** $m < Max$ **do**
  5:    $q_i \leftarrow$ Generate New Offspring $(Q_t)$
  6:    $Q_{t+1} \leftarrow$ Non-Descending Reduce$(Q_t, q_i)$
  7:    $t \leftarrow t + 1$
  8:    $m \leftarrow m + 1$
  9: **end while**

---

In Algorithm 4, the population is firstly divided into non-redundancy part $Q_{nr}$ and redundancy part $Q_r$. If the redundancy set $Q_r$ is not empty, an individual can be selected randomly to be deleted from the population. If there is no individual in $Q_r$, all of the solutions are of non-redundancy type, then 3DCH based sorting without redundancy can be used to rank the population into several priority levels. If there is only one level of solutions, it means that all solutions in the population are non-dominated, then the contribution of each solution to VUS should be calculated and the least contribution one will be deleted from the population. If there are several levels of the population, only the contribution to VUS of individuals on the last priority level should be calculated and the individual with least contribution should be removed from the population.

---

**Algorithm 4** Non-Descending Reduce $(Q, q)$

---

**Require:**     $Q \neq \varnothing$
      $Q$ is a set of solutions
      $q$ is a solution
**Ensure:**     Reduce
1: Split $Q \cup \{q\}$ into two subpopulation $Q_r$ and $Q_{nr}$ ($Q_r$ is the collection of redundancy individuals and $Q_{nr}$ is the collection of non-redundancy individuals)
2: **if** $sizeof(Q_r) >= 1$ **then**
3:     $p \leftarrow$ Randomly select an individual from $Q_r$
4:     $Q \leftarrow Q_{nr} \cup Q_r - \{p\}$
5: **else**
6:     $F_1, ..., F_v \leftarrow$ 3DCH based sorting without redundancy$(Q_{nr})$
7:     **if** $sizeof(F_2) = 0$ **then**
8:         $Vol_{ori} = VUS(Q)$
9:         $Vol_q = VUS(Q + \{q\})$
10:        **if** $Vol_{ori} < Vol_q$ **then**
11:            $d \leftarrow$ The minimum $\Delta$VUS$(F_v)$
12:            $Q \leftarrow Q_{nr} - \{d\}$
13:        **else**
14:            $Q \leftarrow Q$
15:        **end if**
16:    **else**
17:        $d \leftarrow$ The minimum $\Delta$VUS$(F_v)$
18:        $Q \leftarrow Q_{nr} - \{d\}$
19:    **end if**
20: **end if**
21: **return** Q

---

## V. Experimental Studies on Artificial Test Functions

In this section, ZEJD and ZED test functions are designed to test the performance of 3DCH-EMOA and several other EMOAs, such as NSGA-II [36], GDE3 [39], SMS-EMOA [38], SPEA2 [40], MOEA/D [41]. To evaluate the performance of these algorithms VUS, gini coefficient and time cost are adopted in this section. By comparing the results of all algorithms we can make a conclusion that the new proposed algorithm has a good performance to deal with 3-D ROCCH maximization problem. The details of the experiments are described next.

### A. ZEJD Problem

In this section three ZEJD (Zhao, Emmerich, Jiao, Deutz) problems are designed to evaluate the performance of several kinds of EMOAs on ROCCH maximization problems. The test problem is a simulation of augmented ROCCH distribution of complexity classifiers, which has several important properties. Firstly, the points (1,0,0), (0,1,0) and (0,0,1) are included in the Pareto front. Secondly, the Pareto front should be below the ROC surface of random guessing classifiers which is described in Fig. 3. Thirdly, all of the solutions are in the space of the unit cube. The objective of this set of test problems is to find the maximum value of the volume under the convex hull surface. The range of variation of each object is in [0,1], the problem of ZEJD1 is defined in Eq. 9.

$$
\begin{aligned}
f_1 &= 1 - \sqrt{2}cos(x_1 * \pi/2)(1-x_3) \\
f_2 &= 1 - \sqrt{2}sin(x_1 * \pi/2)cos(x_2 * \pi/2)(1-x_3) \\
f_3 &= 1 - \sqrt{2}sin(x_1 * \pi/2)sin(x_2 * \pi/2)(1-x_3)
\end{aligned}
\tag{9}
$$

where $x_1$, $x_2$, $x_3$ are all in $[0,1]$ and $f_1$, $f_2$, $f_3$ are all in $[0,1]$.

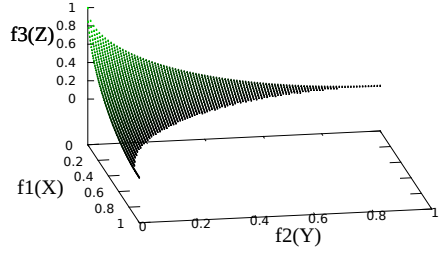

Fig. 7. The Pareto front of ZEJD1 problem

The Pareto front of ZEJD1 is shown in Fig. 7. The problem ZEJD2 and ZEJD3 both are version of ZEJD1 modified by additional dent on the surface. in which some parts of Pareto Front are not on the convex hull. These two test problems are designed to test whether the algorithms can avoid the dent areas. ZEJD2 is defined by Eq. 10, a dent is made in the area satisfied $f_1 < a, f_2 < a, g < a$, by making the function decrease slowly. In our experiments we set $a = 0.3, \lambda = 0.5$, the Pareto front of ZEJD2 is shown in Fig. 8. ZEJD3 is defined by Eq. 11, a dent is made by adding a surface d(x,y) which is shown in Fig. 9. In order to keep the points (1,0,0), (0,1,0) and (0,0,1) in the Pareto front, $d(0,0)$ is subtracted to obtain $f_3$. In this paper, we set $A = 0.15, \gamma = 400$, the Pareto front of ZEJD3 is shown in Fig. 9. The objectives of both ZEJD2 and ZEJD3 are $f_1, f_2$ and $f_3$, $f_1 \in [0,1]$, $f_2 \in [0,1]$, $f_3 \in [0,1]$.

$$
\begin{aligned}
f_1 &= 1 - \sqrt{2}cos(x_1 * \pi/2)(1-x_3) \\
f_2 &= 1 - \sqrt{2}sin(x_1 * \pi/2)cos(x_2 * \pi/2)(1-x_3) \\
g &= 1 - \sqrt{2}sin(x_1 * \pi/2)sin(x_2 * \pi/2)(1-x_3) \\
f_3 &= \begin{cases} a + \lambda(g-a) & \text{if } f_1 < a, f_2 < a, g < a \\ g & \text{else} \end{cases}
\end{aligned}
\tag{10}
$$

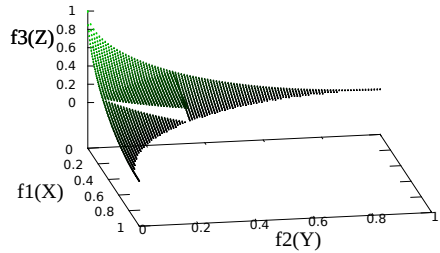

Fig. 8. The Pareto front of ZEJD2 problem

$$f_1 = 1 - \sqrt{2}cos(x_1 * \pi/2)(1 - x_3)$$
$$f_2 = 1 - \sqrt{2}sin(x_1 * \pi/2)cos(x_2 * \pi/2)(1 - x_3)$$
$$g = 1 - \sqrt{2}sin(x_1 * \pi/2)sin(x_2 * \pi/2)(1 - x_3)$$
$$d(x,y) = A * e^{-\gamma\{(x-0.173)^2 + (y-0.173)^2\}}$$
$$k(f_1, f_2) = g + d(f_1, f_2) - d(0,0)$$
$$f_3 = \begin{cases} k(f_1, f_2) & \text{if } k(f_1, f_2) > 0 \\ 0 & \text{else} \end{cases}$$

(11)

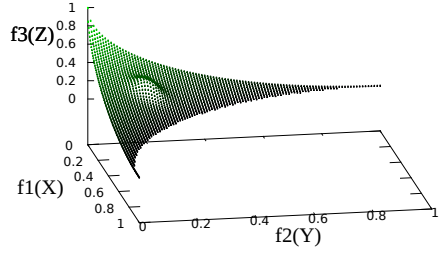

Fig. 9. The Pareto front of ZEJD3 problem

## B. ZED Problem

In this section three ZED (Zhao, Emmerich, Deutz) problems are designed to evaluate the performance of several kinds of EMOAs on ROCCH maximization problems. The test problem is a simulation of 3-D ROC distribution of three-class classifiers, which has several important properties. Firstly, the points (1,0,0), (0,1,0) and (0,0,1) are included in the Pareto front. Secondly, the Pareto front should be above the surface of random guess which is shown in Fig. 5. Thirdly, all of the solutions are in the space of the unit cube. The objective of this set of test problems is to find the maximum value of the volume under the convex hull. The range of variation of each objective is in [0,1], the problem of ZED1 is defined in Eq. 12.

$$f_1 = cos(x_1 * \pi/2)(1 - x_3)$$
$$f_2 = sin(x_1 * \pi/2)cos(x_2 * \pi/2)(1 - x_3)$$
$$f_3 = sin(x_1 * \pi/2)sin(x_2 * \pi/2)(1 - x_3)$$

(12)

where $x_1 \in [0,1]$, $x_2 \in [0,1]$, $x_3 \in [0,1]$ and $f_1 \in [0,1]$, $f_2 \in [0,1]$, $f_3 \in [0,1]$.

The Pareto front of ZED1 is shown in Fig. 10. The problem ZED2 and ZED3 are version of ZED1 modified by additional dent on the surface, in which some parts of Pareto Front are not on the convex hull. These two test problems are designed to test whether the algorithms can avoid the dent areas. ZED2 is defined by Eq. 13, a dent is made in the area satisfied $f_1 > a, f_2 > a, g > a$, by making the function increase slowly. In our experiments we set $a = 0.4, \lambda = 0.5$, the Pareto front of ZED2 is shown in Fig. 11. ZED3 is defined by Eq. 14, a dent is made by subtracting a surface $d(x,y)$ which is shown in Fig. 12. In order to keep the points (1,0,0), (0,1,0) and (0,0,1) in the Pareto front, $d(0,0)$ is added to $k(f_1, f_2)$ as defined in Eq. 14. The points which are less than zero in $k(0,0)$ are set to zero when computing the value to $f_3$. In this paper, we set $A = 0.15, \gamma = 100$, the Pareto front of ZED3 is shown in Fig. 12. The objectives of both ZED2 and ZED3 are $f_1, f_2$ and $f_3$, all of the objectives are in [0,1].

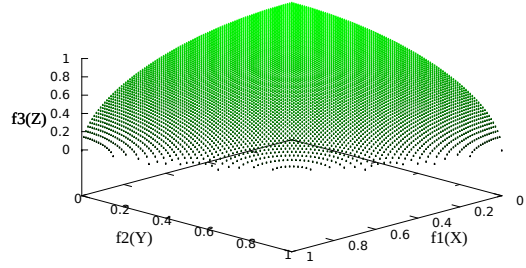

Fig. 10. The Pareto front of ZED1 problem

$$f_1 = cos(x_1 * \pi/2)(1 - x_3)$$
$$f_2 = sin(x_1 * \pi/2)cos(x_2 * \pi/2)(1 - x_3)$$
$$g = sin(x_1 * \pi/2)sin(x_2 * \pi/2)(1 - x_3)$$
$$f_3 = \begin{cases} a + \lambda(g - a) & \text{if } f_1 > a, f_2 > a, g > a \\ g & \text{else} \end{cases} \qquad (13)$$

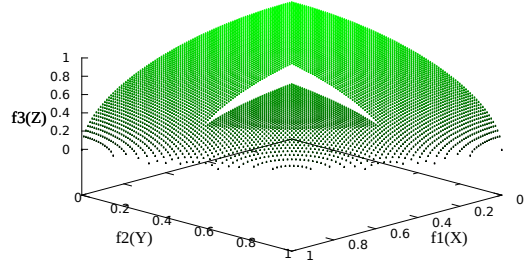

Fig. 11. The Pareto front of ZED2 problem

$$f_1 = cos(x_1 * \pi/2)(1 - x_3)$$
$$f_2 = sin(x_1 * \pi/2)cos(x_2 * \pi/2)(1 - x_3)$$
$$g = sin(x_1 * \pi/2)sin(x_2 * \pi/2)(1 - x_3)$$
$$d(x, y) = A * e^{-\gamma\{(x-0.5)^2 + (y-0.5)^2\}}$$
$$k(f_1, f_2) = g - d(f_1, f_2) + d(0, 0)$$
$$f_3 = \begin{cases} k(f_1, f_2) & \text{if } k(f_1, f_2) > 0 \\ 0 & \text{else} \end{cases} \qquad (14)$$

### C. Metrics

Three metrics are chosen to evaluate the performance of the different algorithms in the comparative experiment on the ZEJD and ZED problems. VUS metric can evaluate the solution set directly, the better the solution set the larger the value of VUS will be. For complicity binary classifiers problem and ZEJD problems the smallest value of VUS is 0 with random guessing classifiers and the largest value of VUS is 0.5. For three-class classification problem and ZED problems the smallest value of VUS is $0$ with random guessing classifiers and the largest value of VUS is $5/6$.

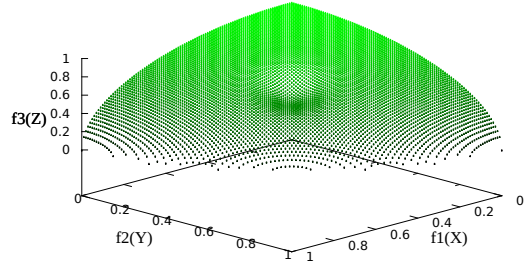

Fig. 12. The Pareto front of ZED3 problem

The gini coefficient is commonly used as a measure of statistical dispersion intended to represent the income distribution of a nation's residents [34]. In this work, the gini coefficient is used to evaluate the uniformity of the solution set by calculating the statistical distribution of the nearest neighbor distance of each solution. The gini coefficient can describe the spread of neighboring individuals on the achieved Pareto front. The value of gini coefficient will be zero if the solution set is distributed uniformly. The definition of gini coefficient is described in Eq. 15. The lower the value of gini coefficient the more evenly distributed the solution set will be.

$$G = \frac{1}{n}\left(n + 1 - 2\left(\frac{\sum_{i=1}^{n}(n+1-i)d_i}{\sum_{i=1}^{n}d_i}\right)\right) \tag{15}$$

where $G$ represents the value of gini coefficient, $n$ is the number of solutions in the set, $d_i$ is the nearest neighbor distance for each solution in the space.

Time cost is used to measure the complexity of each algorithm in this section. In general, the most complex algorithms are most computationally expensive.

*1) Parameters Setting:* All algorithms are run for 25000 function evaluations. The simulated binary crossover (SBX) operator and the polynomial mutation are applied in all experiments. The crossover probability of $p_c = 0.9$ and a mutation probability of $p_m = 1/n$ where n is the number of decision variables are used. The population size is set to 50 for ZEJD problem and 100 for ZED problem, because the convex hull surface of ZED is larger than the convex hull surface of ZEJD. The size of archive for SPEA2 is equal to the size of the population. All of the experiments are based on jMetal framework [42], [43]. All of the algorithms are run 30 times independently.

### D. Experimental results and discussions

*1) ZEJD problem:* The comparison of simulation experiments with NSGA-II, GDE3, SPEA2, MOEA/D, SMS-EMOA, and 3DCH-EMOA on ZEJD problems is discussed in this section. The results of experiments are given as follows: the results not only include the plots of solution set in the objective space but also include statistical analysis on the metrics of these results. The illustrations of solutions set in the $f_1 - f_2 - f_3$ objectives space are plotted for the ZEJD problems. The solutions obtained are depicted in dark dots and the true non-dominated solutions are in green dots.

The results of ZEJD1 are shown in Fig. 13, of ZEJD2 are shown in Fig. 14 and of ZEJD3 are shown in Fig. 15. By comparing the Pareto fronts of all the results we can make some conclusions: NSGA-II, GDE3 and SPEA2 show the worst convergence. MOEA/D can converge to the true Pareto front, however it does not give good results on diversity and distribution uniformity. The result of MOEA/D has no solutions on the edges of the solution space. SMS-EMOA and 3DCH-EMOA have good performance on convergence, diversity and distribution uniformity. However the points near (0,0,1) are not included in the results of SMS-EMOA, which results in bad performance of VUS metric. There are several points on the dent areas in the results on ZEJD2 and ZEJD3 for the most of the algorithms. However, the new proposed algorithm 3DCH-EMOA can always heal the dent area of ZEJD2 and ZEJD3 test problems, but other algorithms are not so reliable. In summary, 3DCH-EMOA can always achieve better results than other algorithms, not only on convergence and distribution uniformity, but also has the ability to heal the dent area.

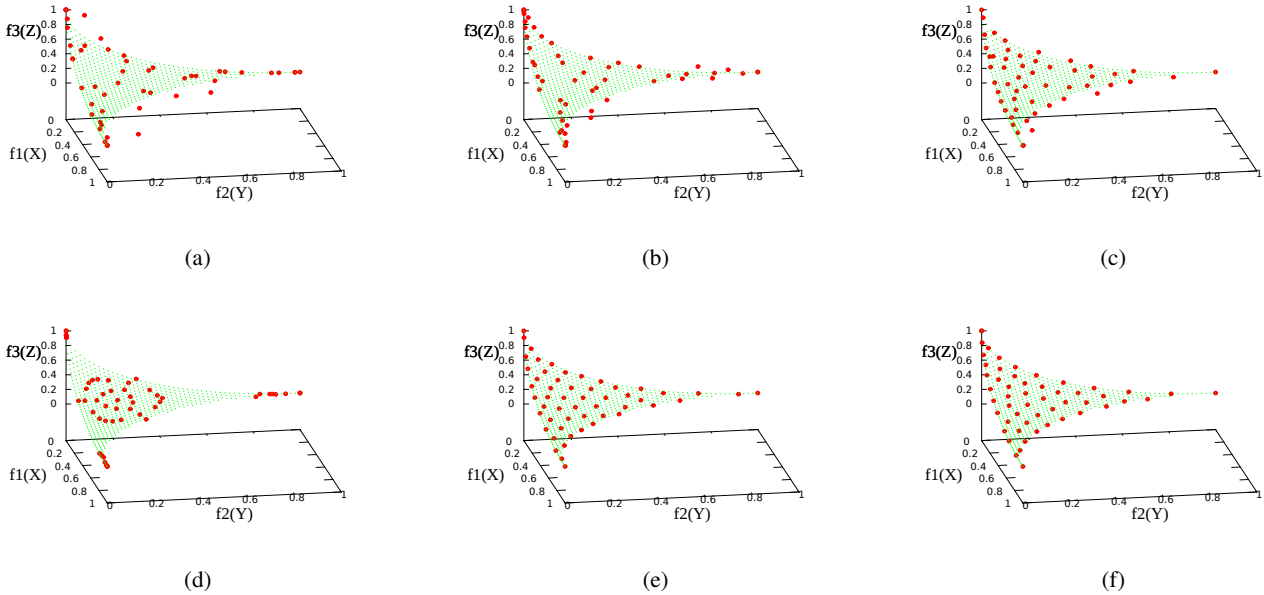

Fig. 13. Experimental results of ZEJD1 are shown in $f_1 - f_2 - f_3$ space. (a) Result of NSGA-II. (b) Result of GDE3. (c) Result of SPEA2. (d) Result of MOEA/D. (e) Result of SMS-EMOA. (f) Result of 3DCH-EMOA.

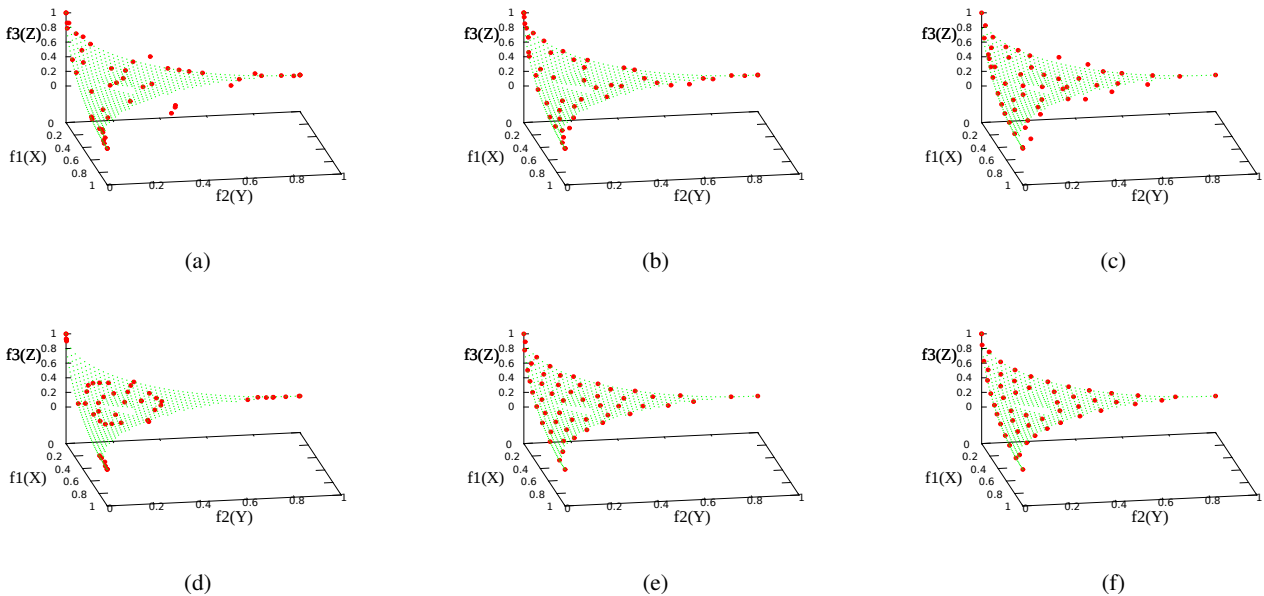

Fig. 14. Experimental results of ZEJD2 are shown in $f_1 - f_2 - f_3$ space. (a) Result of NSGA-II. (b) Result of GDE3. (c) Result of SPEA2. (d) Result of MOEA/D. (e) Result of SMS-EMOA. (f) Result of 3DCH-EMOA.

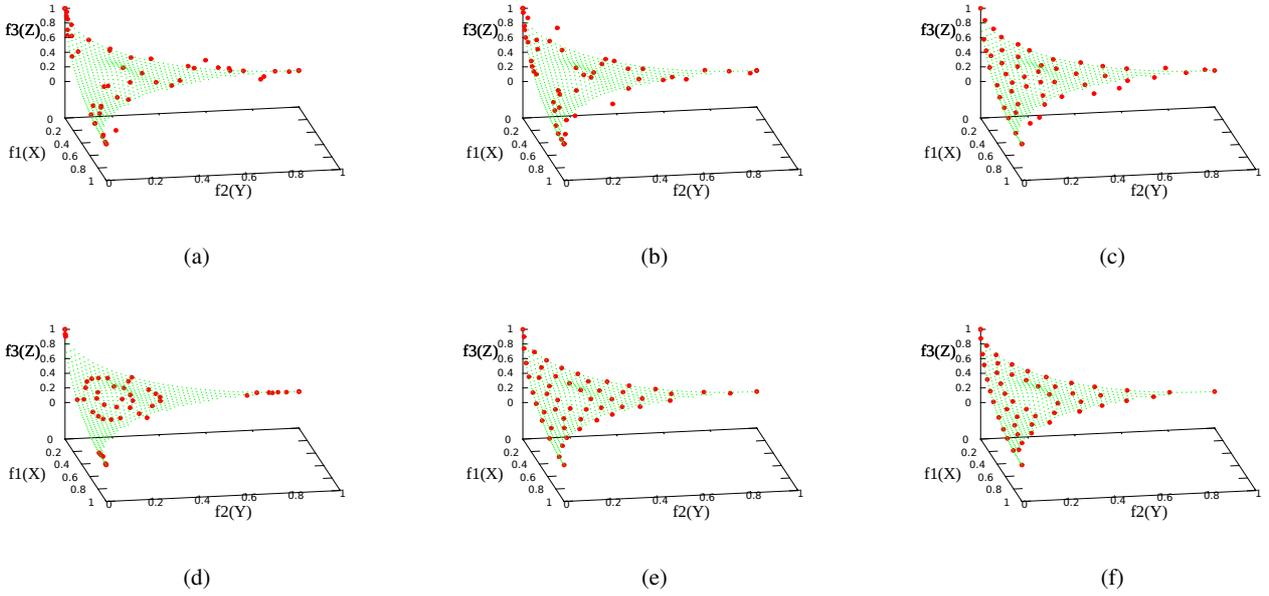

Fig. 15. Experimental results of ZEJD3 are shown in $f_1 - f_2 - f_3$ space. (a) Result of NSGA-II. (b) Result of GDE3. (c) Result of SPEA2. (d) Result of MOEA/D. (e) Result of SMS-EMOA. (f) Result of 3DCH-EMOA.

TABLE I
VUS. MEAN AND STANDARD DEVIATION

|  | NSGA-II | GDE3 | SPEA2 | MOEA/D | SMS-EMOA | 3DCH-EMOA |
|---|---|---|---|---|---|---|
| ZEJD1 | $4.60e-01_{1.3e-03}$ | $4.62e-01_{6.3e-04}$ | $4.49e-01_{1.1e-02}$ | $4.59e-01_{2.1e-03}$ | $4.46e-01_{3.6e-03}$ | $4.65e-01_{5.0e-06}$ |
| ZEJD2 | $4.60e-01_{1.2e-03}$ | $4.61e-01_{6.4e-04}$ | $4.48e-01_{1.3e-03}$ | $4.59e-01_{1.3e-03}$ | $4.46e-01_{3.1e-03}$ | $4.64e-01_{4.5e-06}$ |
| ZEJD3 | $4.60e-01_{8.8e-04}$ | $4.61e-01_{6.9e-04}$ | $4.46e-01_{1.2e-02}$ | $4.59e-01_{1.5e-03}$ | $4.46e-01_{4.1e-03}$ | $4.64e-01_{3.9e-06}$ |

In the experiments, each algorithm is run for 30 times independently on ZEJD test problems to evaluate and compare the robustness of these algorithms. The performance characteristics of each algorithm can be seen from the statistical analysis of the experimental results. The statistical results (mean and standard deviation) of the VUS are shown in Table.I. The detailed discussion follows next.

While dealing with the ZEJD problems and considering the metric of VUS, 3DCH-EMOA gets the largest value of mean and the smallest value of standard deviation, which shows that 3DCH-EMOA has a good performance not only in convergence but also in stability. GDE3 obtains the second best result with these test functions. SPEA2 and SMS-EMOA did not have good performance because they missed points near (0,0,1) in the solution space.

The statistical results of the gini coefficient are shown in Table.II. By comparing the results on the table we can see that 3DCH-EMOA gets the smallest value of mean and the smallest value of standard deviation, which shows that 3DCH-EMOA has a good performance in uniformity and diversity of the population. SPEA2 obtains the second best result, however it did not have good performance on convergence. SMS-EMOA obtains the third best result and it also has good performance on convergence.

The statistical results of time cost of optimization are shown in Table. III. MOEA/D always cost the least time and NSGA-II performs better than others. SMS-EMOA is the most time consuming algorithm and 3DCH-EMOA only performs better than SMS-EMOA. In the case of machine learning problems, such as feature selection and parameters optimization of classifiers, the evaluation takes much more time than optimization process, which is different from test functions. Considering problems in machine learning, the cost of time of optimization is not a key obstacle, especially for off line learning area, 3DCH-EMOA plays well with it which is shown in the next section.

TABLE II
GINI COEFFICIENT. MEAN AND STANDARD DEVIATION

|       | NSGA-II | GDE3 | SPEA2 | MOEA/D | SMS-EMOA | 3DCH-EMOA |
|-------|---------|------|-------|--------|----------|-----------|
| ZEJD1 | $3.83e-01_{4.0e-02}$ | $2.90e-01_{2.9e-02}$ | $9.60e-02_{2.0e-02}$ | $3.09e-01_{1.7e-02}$ | $9.88e-02_{1.3e-02}$ | $8.18e-02_{1.4e-02}$ |
| ZEJD2 | $3.58e-01_{3.9e-02}$ | $2.87e-01_{2.7e-02}$ | $9.49e-02_{1.5e-02}$ | $3.69e-01_{2.6e-02}$ | $9.54e-02_{1.4e-02}$ | $8.74e-02_{1.6e-02}$ |
| ZEJD3 | $3.45e-01_{4.5e-02}$ | $2.92e-01_{2.3e-02}$ | $9.10e-02_{1.6e-02}$ | $3.12e-01_{1.6e-02}$ | $9.62e-02_{1.4e-02}$ | $8.67e-02_{1.2e-02}$ |

TABLE III
TIME COST OF OPTIMIZATION. MEAN AND STANDARD DEVIATION

|       | NSGA-II | GDE3 | SPEA2 | MOEA/D | SMS-EMOA | 3DCH-EMOA |
|-------|---------|------|-------|--------|----------|-----------|
| ZEJD1 | $1.29e+02_{9.6e+01}$ | $2.91e+03_{2.2e+01}$ | $2.05e+03_{6.9e+01}$ | $8.49e+01_{6.5e+01}$ | $7.03e+04_{2.5e+03}$ | $5.30e+04_{1.4e+03}$ |
| ZEJD2 | $1.32e+02_{1.0e+02}$ | $2.91e+03_{3.8e+01}$ | $2.04e+03_{6.5e+01}$ | $8.82e+01_{6.6e+01}$ | $6.80e+04_{2.3e+03}$ | $4.41e+04_{1.0e+03}$ |
| ZEJD3 | $1.25e+02_{7.8e+01}$ | $2.61e+03_{2.8e+02}$ | $2.07e+03_{1.9e+01}$ | $8.08e+01_{3.8e+01}$ | $7.33e+04_{3.6e+03}$ | $4.91e+04_{1.3e+03}$ |

*2) ZED problem:* The comparison of simulation experiments with NSGA-II, GDE3, SPEA2, MOEA/D, SMS-EMOA, and 3DCH-EMOA on ZED problems is discussed in this section. The results of experiments are given in the following, not only the plots of the solution sets are included but also the statistical results on the metrics are analyzed. The illustrations of solution set are plotted in the $f_1 - f_2 - f_3$ objectives space, in which the solutions obtained by each algorithm are depicted in dark dots and the true non-dominated solutions in green dots.

The results of ZED1 are shown in Fig. 16, of ZED2 are shown in Fig. 17 and of ZED3 are shown in Fig. 18. By comparing the Pareto fronts of all the results we can make some conclusions: NSGA-II and GDE3 show the worst convergence. MOEA/D can converge to the true Pareto front, however it does not give good result on diversity and distribution uniformity. The result of MOEA/D has too many solutions on the edges and the middle area of solution space. SPEA2, SMS-EMOA and 3DCH-EMOA have good performance on convergence, diversity and distribution uniformity. However there are no solutions on the edges of solution space for SMS-EMOA, and the distribution of solutions on the edges of solution space for SPEA2 are not as uniform as with 3DCH-EMOA. There are several points on the dent areas for the results of ZED2 and ZED3 problems for the most of the algorithms. However, the new proposed algorithm can always heal the dent area of ZED2 and ZED3 test problems, but the other algorithms can not. In summary, 3DCH-EMOA can always achieve better results than other algorithms, not only on convergence and distribution uniformity but also has the ability to heal the dent area.

TABLE IV
VUS. MEAN AND STANDARD DEVIATION

|      | NSGA-II | GDE3 | SPEA2 | MOEA/D | SMS-EMOA | 3DCH-EMOA |
|------|---------|------|-------|--------|----------|-----------|
| ZED1 | $3.36e-01_{4.7e-03}$ | $3.43e-01_{2.7e-03}$ | $3.49e-01_{6.1e-04}$ | $3.43e-01_{2.3e-04}$ | $3.33e-01_{6.9e-04}$ | $3.53e-01_{1.3e-05}$ |
| ZED2 | $3.33e-01_{3.5e-03}$ | $3.40e-01_{1.7e-03}$ | $3.47e-01_{8.5e-04}$ | $3.42e-01_{3.1e-04}$ | $3.32e-01_{8.2e-04}$ | $3.52e-01_{1.7e-05}$ |
| ZED3 | $3.36e-01_{3.1e-03}$ | $3.40e-01_{1.7e-03}$ | $3.47e-01_{7.3e-04}$ | $3.41e-01_{2.5e-04}$ | $3.31e-01_{8.1e-04}$ | $3.51e-01_{2.0e-05}$ |

In these experiments, each algorithm is run 30 times independently on ZED test problems to evaluate and compare the robustness of the algorithms. The performance characteristics of each algorithm can be seen from the statistical analysis of the experimental results. The statistical results of the VUS are shown in Table.IV. The detailed discussion follows next.

For the ZED1, ZED2 and ZED3 problem, 3DCH-EMOA gets the largest value of mean and the smallest value of standard deviation of VUS, which shows that 3DCH-EMOA has good performance not only in convergence but also in stability. SPEA2 obtains better results than the remaining algorithms with these test functions.

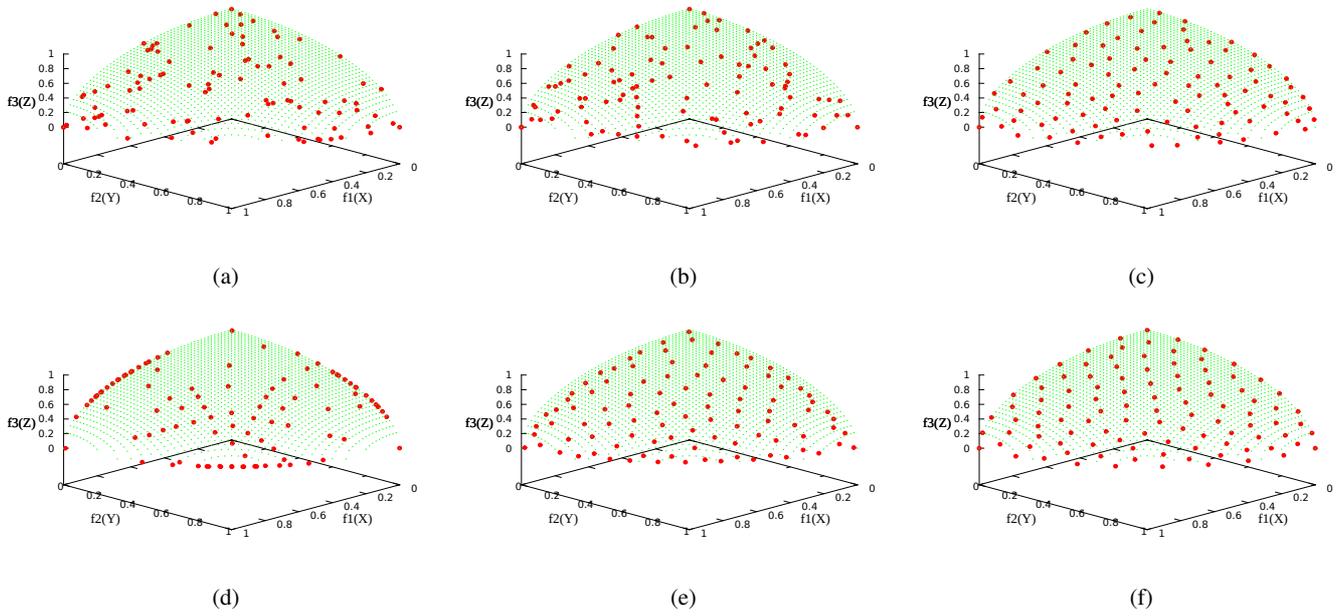

Fig. 16. Experimental results of ZED1 are shown in $f_1 - f_2 - f_3$ space. (a) Result of NSGA-II. (b) Result of GDE3. (c) Result of SPEA2. (d) Result of MOEA/D. (e) Result of SMS-EMOA. (f) Result of 3DCH-EMOA.

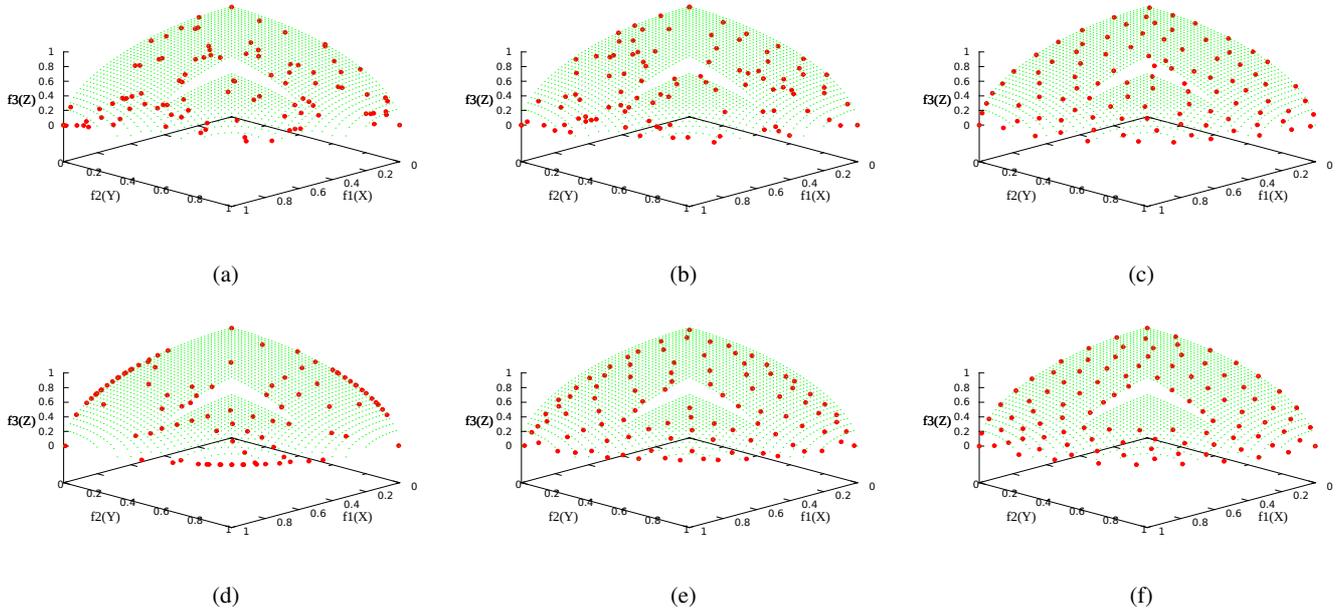

Fig. 17. Experimental results of ZED2 are shown in $f_1 - f_2 - f_3$ space. (a) Result of NSGA-II. (b) Result of GDE3. (c) Result of SPEA2. (d) Result of MOEA/D. (e) Result of SMS-EMOA. (f) Result of 3DCH-EMOA.

The statistical results of the gini coefficient are shown in Table.V. The detailed discussion of each problem follows next.

For the ZED1, ZED2 and ZED3 problem, 3DCH-EMOA gets the smallest value of mean and the smallest value of standard deviation of the gini coefficient, which shows that 3DCH-EMOA has a good performance in uniformity. SPEA2 obtains better result than others but 3DCH-EMOA.

The statistical results of time cost of optimization are shown in Table. VI. NSGA-II always computa-tionally least expensive and MOEA/D performs better than others but NSGA-II. SMS-EMOA is the most time cost algorithm and 3DCH-EMOA only performs better than SMS-EMOA.

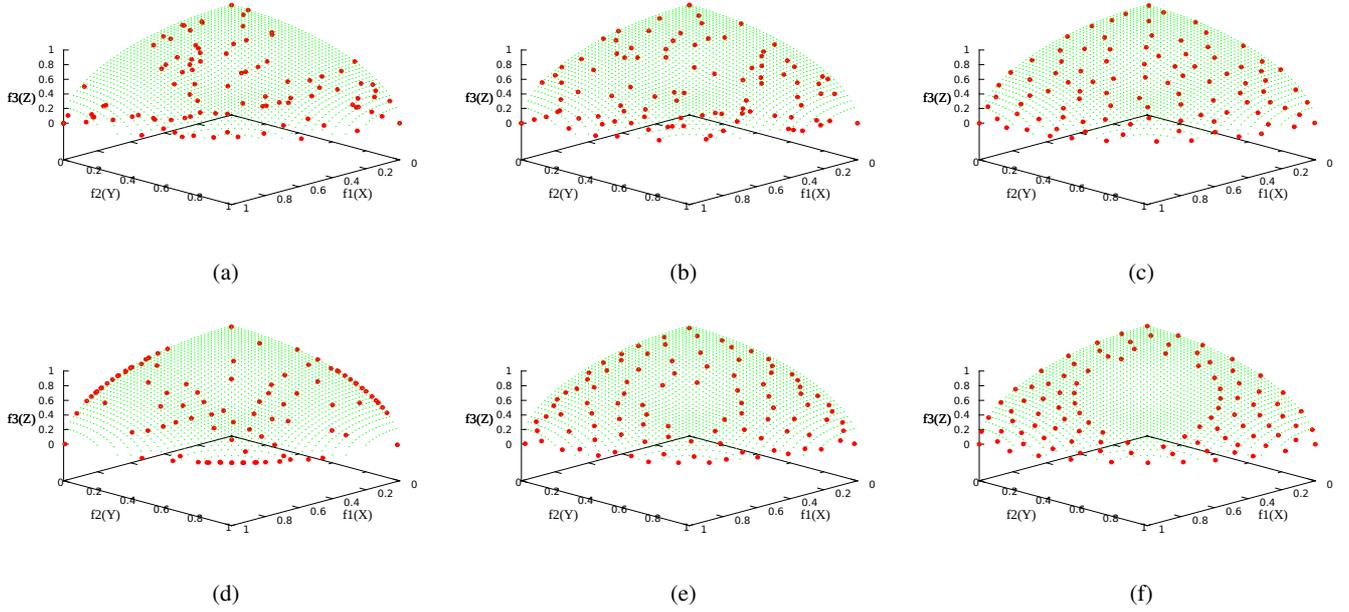

Fig. 18. Experimental results of ZED3 are shown in $f_1 - f_2 - f_3$ space. (a) Result of NSGA-II. (b) Result of GDE3. (c) Result of SPEA2. (d) Result of MOEA/D. (e) Result of SMS-EMOA. (f) Result of 3DCH-EMOA.

TABLE V
Gini Coefficient. Mean and standard deviation

| | NSGA-II | GDE3 | SPEA2 | MOEA/D | SMS-EMOA | 3DCH-EMOA |
|---|---|---|---|---|---|---|
| ZED1 | $3.39e-01_{3.5e-02}$ | $2.74e-01_{2.1e-02}$ | $6.60e-02_{5.3e-03}$ | $4.51e-01_{4.4e-03}$ | $1.16e-01_{6.8e-03}$ | $4.64e-02_{4.4e-03}$ |
| ZED2 | $3.35e-01_{3.1e-02}$ | $2.77e-01_{2.0e-02}$ | $6.53e-02_{6.6e-03}$ | $4.62e-01_{7.3e-03}$ | $1.12e-01_{5.7e-03}$ | $5.37e-02_{5.5e-03}$ |
| ZED3 | $3.40e-01_{2.8e-02}$ | $2.69e-01_{2.0e-02}$ | $6.69e-02_{7.1e-03}$ | $4.49e-01_{6.1e-03}$ | $1.10e-01_{6.3e-03}$ | $6.12e-02_{5.7e-03}$ |

The new proposed algorithm is computationally expensive due to large computational resources needed for computing the VUS contribution of every point in the solution space. The dynamic convex hulls algorithm [44] will be adapted in future work to reduce the computational complexity of the new method.

TABLE VI
Time cost of optimization. Mean and standard deviation

| | NSGA-II | GDE3 | SPEA2 | MOEA/D | SMS-EMOA | 3DCH-EMOA |
|---|---|---|---|---|---|---|
| ZED1 | $1.73e+02_{3.9e+01}$ | $1.26e+04_{3.7e+02}$ | $4.60e+03_{1.6e+02}$ | $8.67e+01_{1.0e+01}$ | $7.19e+05_{2.2e+04}$ | $2.89e+05_{5.5e+03}$ |
| ZED2 | $1.71e+02_{2.3e+01}$ | $1.28e+04_{4.7e+02}$ | $4.56e+03_{6.5e+01}$ | $8.40e+01_{4.3e+00}$ | $7.22e+05_{2.3e+04}$ | $2.71e+05_{4.4e+03}$ |
| ZED3 | $1.86e+02_{8.0e+01}$ | $1.26e+04_{2.9e+02}$ | $4.73e+03_{2.4e+02}$ | $9.51e+01_{2.7e+01}$ | $7.48e+05_{2.5e+04}$ | $2.96e+05_{3.2e+03}$ |

## VI. SPAM problem

From a technical point of view, an email anti-SPAM system consists of a set of boolean filtering rules, that jointly allows for SPAM messages detection. Discovering the relative importance of these rules and assigning the corresponding scores (weights) of each rule, is a complex setup process.

The need of frequent scores reassignment for existing rules and setting scores for new rules, to keep the anti-SPAM filter updated and running, requires the adoption of machine learning and optimization techniques.

### A. SPAM multiobjective problem formulation

SPAM filtering problem optimization has been addressed by the techniques surveyed in [45], [46]. The formulation of the scores setting optimization problem is naturally bi-objective, a typical user would wish to minimize both, the number of SPAM messages not identified by anti-spam filtering techniques, called false negative rate (fnr), and the number of legitimate messages classified as SPAM by mistake, called false positive rate (fpr). A business email is one of extreme cases of anti-SPAM systems setup with such objectives, where the fpr and fnr should be tuned to have lowest possible rate of legitimate messages lost, usually at the expenses of higher false negative classifications. On the other extreme is content management systems (CMSs) devoted to entertainment, where dismissing some legitimate messages keeps or improves the interest on their usage, while the acceptance of any SPAM message is not allowed. The cases between these two extremes are also of high interest for a variety of the problem areas.

In previous work on anti-SPAM filter optimization [46] it was observed that many rules were not participating in the classification process and some (with very small weights) only marginally influenced the classification results. This observation suggests that in addition to optimizing fpr and fnr, the complexity of the anti-SPAM filter or its parsimony can be optimized.

The trend of increasing the number of rules for the system operation creates an empirically known potential inefficiency phenomenon, which is addressed under the so called principle of parsimony. This principle states, in one of its simplified formulations, that unnecessary assumptions for a theory (conclusion) should not be considered if they have no consequence for the conclusion [47].

Parsimony is measured in the context of the current anti-SPAM study, as the minimum number of rules with score different from zero, that support a specific classification quality.

In our study we also follow a three-objective problem formulation, minimizing all three objectives, fpr, fnr and ccr (number of anti-SPAM filter rules rate) to be used in the classification process.

*1) SpamAssassin Corpus:* For the multiobjective anti-SPAM problem formulation experiments, we adopted the SpamAssassin system [48]. SpamAssassin was selected due to its popularity and wide adoption by the open source community, the research community on anti-SPAM systems, wide commercial usage, and available email corpora. The SpamAssassin corpus used in our experiments is composed of 9349 email messages, 2398 SPAM and 6951 legitimate messages [49].

SpamAssassin became a reference in the anti-SPAM filtering domain, not only due to its public availability to research and development, but also because of its performance (classification quality).

*2) Algorithms Involved:* Five reference multiobjective algorithms were tested (NSGA-II, SPEA2, MOEA/D, SMS-EMOA and 3DCH-EMOA), for the three objectives anti-SPAM filtering problem formulation, using the SpamAssassin corpus [49] for SPAM classification quality assessment. Experiments were performed with jMetal [43], an optimization framework for the development of multiobjective metaheuristics in Java.

*3) Configuration and Evaluation:* Default parameters for problem formulation, experiments and algorithms settings were adopted for the experiments.

Encoding: We employed a jMetal RealBinary encoding scheme where the chromosome is constituted by an array of real values in the interval $[-5; 5]$ and a bit string of equal length. The length of the chromosome is determined by the number of anti-SPAM filtering rules. In this study the rules available in the SpamAssassin software public distributions that effectively match SpamAssassin email messages corpus is 330. Each rule is associated with a real value score in the $[-5, 5]$ interval and a one bit in the chromosome. If the $i$th bit is 0 the $i$th rule is ignored, and otherwise the rule is considered by the SPAM classifier with the $i$th corresponding real value score (weight). Messages are classified as SPAM when the sum of the active rules that match the message is equal or greater than the threshold value of 5.

Configuration: The five (NSGA-II, SPEA2, SMS-EMOA, MOEA/D, 3DCH-EMOA) algorithms are set with a maximum of 25000 function evaluations as the experiment stopping criteria. The SBX single point crossover and polynomial bit flip mutation operators are applied in the experiments. The crossover probability of $p_c = 0.9$ and a mutation probability of $p_m = 1/n$ where n is the number of anti-spam filtering rules. The population size is set to 100 for all algorithms, archive size 100 for SPEA2 and offset 100 for SMS-EMOA and 3DCH-EMOA. All of the algorithms are run 30 times independently.

*4) Results and analysis:* The comparison of NSGA-II, MOEA/D, SPEA2, SMS-EMOA and 3DCH-EMOA algorithms for the three objectives SPAM problem formulation is done with respect to the reference Pareto front, which is taken as a close approximation of the true Pareto front. The reference Pareto front is calculated as the best set of solutions of all algorithms achieved in all experimental runs.

We will first interpret this reference Pareto front shown in Fig. 19, Fig. 20 and Fig. 21, corresponding to ccr x fpr (classifier complexity ratio x false positive rate), ccr x fnr (classifier complexity ratio x false negative rate) and the three axis projections, respectively (all objectives are to be minimized).

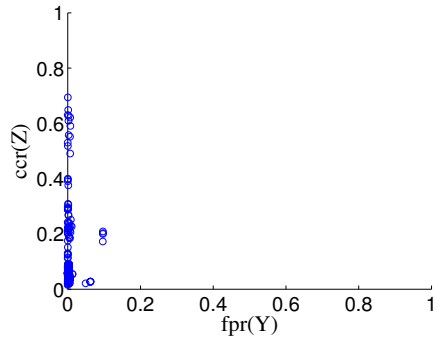

Fig. 19.  Reference Pareto front for three objectives SPAM problem formulation (ccr x fpr projection)

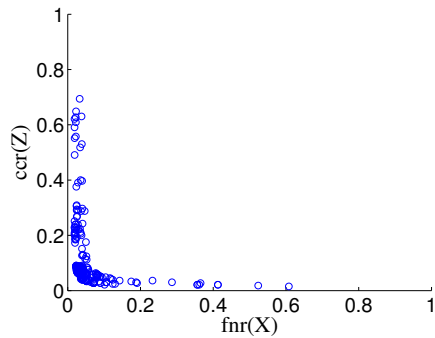

Fig. 20.  Reference Pareto front for three objectives SPAM problem formulation (ccr x fnr projection)

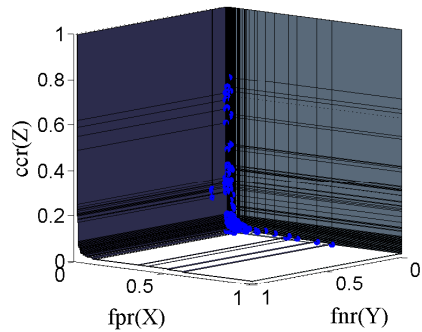

Fig. 21.  Reference Pareto front for three objectives SPAM problem formulation (three axis projection)

The plots show the boundary between the dominated and non-dominated space (attainment curve). All values are percentages relative to the number of anti-SPAM filtering rules (330) and number of email messages (9349).

From the plots we conclude that: 1) Even for a classifier using the maximal number of rules, the fnr could not be reduced to zero, but it got very close to it; 2) The fpr is almost exactly zero for SPAM filters that use only ca. 15% of the rules; 3) Using about 20% of the rules, the knee point solution is found. From then on, only marginal improvements are possible by adding more rules.

In summary, the adding of a third objective is particularly valuable because it can help to reduce the computational effort for the classification to ca. 20% of the effort when all rules are used, losing almost no performance. The second question is how close different algorithms get to the true Pareto front, here represented by the reference Pareto front. For this, one might look at Pareto fronts of each algorithm that have an average performance in VUS. Also, we can look at summary statistics on performance metrics, first and foremost on the VUS performance.

The performance statistics in Table VII, Fig. 22 and Fig. 23 indicate that the 3DCH-EMOA has clearly the best performance in the VUS metric and also it achieves a relative good SPREAD. Because the VUS metric is the most relevant to 3-D ROC optimization, it is recommended to use 3DCH-EMOA for finding Pareto front approximations for the 3-D anti-SPAM filtering problem. Interestingly, the results on the hypervolume metric, which is also a volume based indicator, are relatively bad for 3DCH-EMOA (Fig. 24). This shows that a good performance in the hypervolume metric does not coincide with a good performance in the VUS metric.

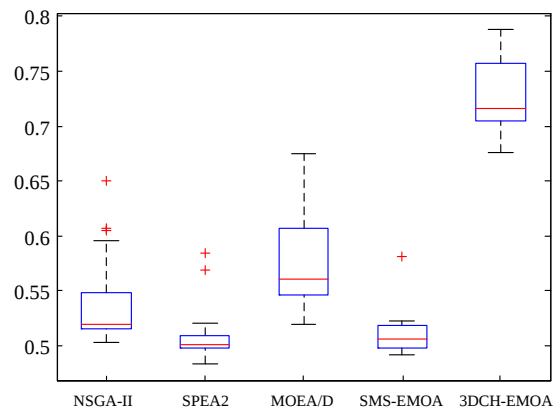

Fig. 22. VUS boxplot for three objectives SPAM problem formulation

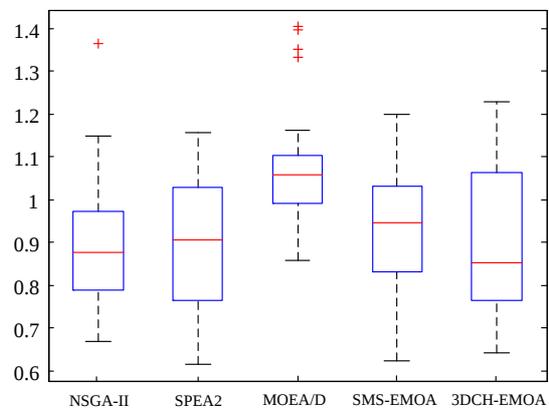

Fig. 23. SPREAD boxplot for three objectives SPAM problem formulation

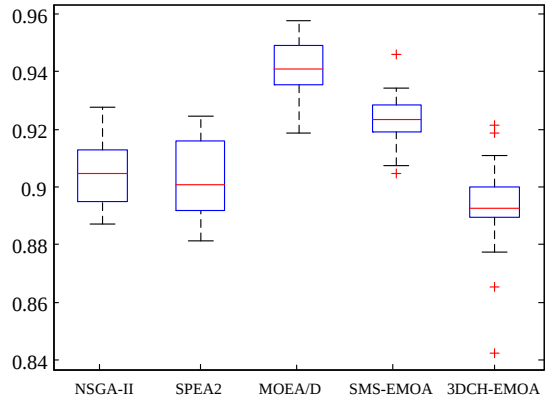

Fig. 24. Hypervolume boxplot for three objectives SPAM problem formulation

TABLE VII
VUS. MEAN AND STANDARD DEVIATION

|  | NSGA-II | MOEA/D | SPEA2 | SMS-EMOA | CHEMOA3D |
|---|---|---|---|---|---|
| SPAM | $3.41e-01_{5.8e-03}$ | $3.48e-01_{1.3e-02}$ | $3.31e-01_{7.8e-03}$ | $3.26e-01_{7.7e-03}$ | $4.08e-01_{4.9e-03}$ |

## VII. FEATURE SELECTION AND CLASSIFICATION

Feature selection is a data mining technique to mine useful information in the vast amount of information. Feature selection methods provide us a way of reducing computation time, improving prediction performance, and a better understanding of the data in machine learning or pattern recognition applications [50].

These methods refer to the process of selecting descriptors that are most effective while reducing effects from noise or irrelevant variables and still provide good prediction results. Feature selection aims at reducing the number of given features to the subset of most useful features. To evaluate all the subsets of set with N feature, $2^N$ evaluations are needed. The problem becomes an NP-hard problem as the number of features grows. Evolutionary algorithms have proved to be particularly useful to discover significant and meaningful information in large quantities of raw noisy data [51].

### A. Binary classification problem

Generally speaking, the more features in a dataset, the more information would be available for classification. However, because of some irrelevant and redundant features in the data set, not all of the features have a positive contribution for the classification process. Actually, reducing the number of features in a data set can lead to faster model training and to improve the performance of classifiers.

The feature selection problem for binary classification $(\Omega, F)$ can formally be defined as a multiobjective optimization problem: determine the feature set $S$ as it is described in Eq. 16.

$$\min F(S) = (fpr, fnr, ccr)$$
$$\text{subject to} \quad S \in \Omega \qquad (16)$$

where $\Omega$ is the set of all possible feature subsets and $S$ refers to feature subsets.

Besides fpr and fnr for binary classifiers, we define the third objective as used features rate (ufr or ccr). The subset of features with lower ccr should be preferred given the same fpr and fnr. The computational cost of a classifier with higher ccr is considered to be higher than that with a lower ccr. It is possible to construct a classifier with different classifiers trained by subsets of features in 3-D ROC space by means of randomization, which has computation complexity between them [1].

*1) Data sets:* Twenty-four data sets are selected from UCI repository [18] and described in Table VIII. Balanced and imbalanced benchmark data sets are included, and the number of instances of these data sets ranges from hundreds to thousands.

TABLE VIII
UCI DATA SETS

| No. | Data Set | No.of features | Class Distribution |
|---|---|---|---|
| 1 | australian | 14 | 383:307 |
| 2 | breast-w | 9 | 458:241 |
| 3 | credit-a | 15 | 307:383 |
| 4 | cylinder-bands | 39 | 228:312 |
| 5 | diabetes | 8 | 500:268 |
| 6 | german | 24 | 700:300 |
| 7 | heart | 13 | 150:120 |
| 8 | hepatitis | 19 | 32:123 |
| 9 | hypothyroid-bi | 29 | 3481:291 |
| 10 | labor | 16 | 20:37 |
| 11 | mushroom | 22 | 4208:3916 |
| 12 | ionosphere | 34 | 126:225 |
| 13 | kr-vs-kp | 36 | 1669:1527 |
| 14 | parkinsons | 22 | 147:48 |
| 15 | pima | 8 | 268:500 |
| 16 | sick | 29 | 3541:231 |
| 17 | sonar | 60 | 97:111 |
| 18 | spambase | 57 | 2788:1813 |
| 19 | spect | 22 | 55:212 |
| 20 | spectf | 44 | 95:254 |
| 21 | tic-tac-toe | 9 | 332:626 |
| 22 | vote | 16 | 267:168 |
| 23 | wdbc | 30 | 212:357 |
| 24 | wpbc | 33 | 151:47 |

*2) Tested algorithms:* To evaluate the performance of the new proposed method, comparisons with state of the art EMOAs such as NSGA-II, SPEA2 and SMS-EMOA were done. For classification performance comparison the C4.5 classifier was adopted. It is an algorithm used to generate a decision tree, proposed in [52], with an open source Java implementation (J48), available in the weka data mining tool [53].

*3) Configuration and Evaluation:* Default parameters for problem formulation, experiments and algorithms settings were adopted.

Encoding: We employed binary encoding scheme where the chromosome is a bit string. The length of the chromosome is determined by the number of features of the dataset. Each feature is associated with one bit in the chromosome. If the $i$th bit is 0 the $i$th feature is ignored, otherwise that feature is selected. Each chromosome represents a subset of features and a set of parameters of the classifier.

Configuration: All algorithms are run for 2000 classification evaluations. The single point crossover operator and bit flip mutation are applied in the experiments. The crossover probability of $p_c = 0.9$ and a mutation probability of $p_m = 1/n$ where n is the number of variables. The population size is set to 20 for all algorithms and the size of SPEA2 archive is the 20 too. We apply each algorithm on each data set with five-fold cross-validation for 10 times independently. All experiments are based on jMetal [43] and Weka [53].

*4) Results and Analysis:* The performance comparison of NSGA-II, SPEA2, SMS-EMOA and 3DCH-EMOA on feature selection with UCI data sets is discussed here. The results include the plots of augmented ROC space and statistical analysis on VUS metric. The Pareto front of the feature selection result of vote data set is shown in Fig. 25 and the details are discussed below.

The reference Pareto front is shown in Fig. 25(e), which is obtained by putting the results of all different algorithms together. Some conclusions can be made from the reference Pareto front. The vote data set can obtain good performance with about 20% to 100% of features. The performance of the classification decreases slowly with the reduction of features from 100% to 20%. With less than 20% of the features a signification decrease in performance occurs. Only the new proposed method can obtain result similar with reference Pareto front. The other algorithms missed the subsets with more than 45% of features. The new proposed algorithm is more useful in reality.

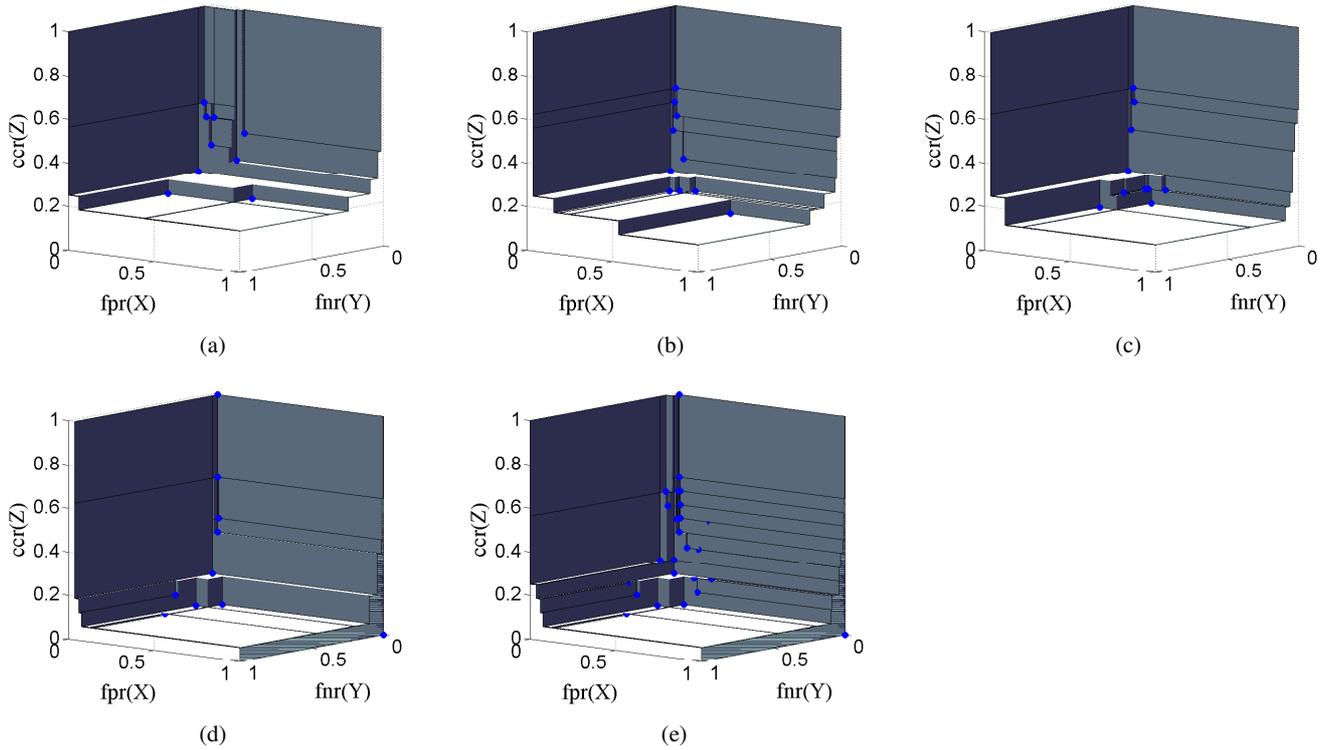

Fig. 25. Experimental results of feature selection of vote data set are shown in $fpr - fnr - ccr$ space. (a) Result of NSGA-II in augmented ROC space. (b) Result of SPEA2. (c) Result of SMS-EMOA. (d) Result of 3DCH-EMOA. (h) Result of the reference Pareto front.

The statistical results (mean and standard deviation) of the VUS metric are shown in Table.IX. For most of the results, 3DCH-EMOA obtains the largest value of mean and small value of standard deviation, which shows that 3DCH-EMOA not only has good performance in convergence but also has good performance in stability. SPEA2 is the second best performing algorithm on most of data sets.

### B. Three-class classification

In this part, we focus on the feature selection for three-class classification problem. With the technique of iso-performance [1] we can obtain any classifiers on the surface of ROCCH. Generally, the larger value of the VUS the more diverse the set of classification will be. Ensemble learning is a powerful learning approach that combines multiple classifiers to build up more accurate classifier than each of the individual classifier [54]. The performance of ensemble learning always rely on the diversity of feature subset selection [55].

For three-class classification problem, we define the three objectives as far, fbr and fcr. The feature selection problem is described in Eq. 17.

$$\max F(S) = (tar, tbr, tcr)$$
$$\text{subject} \quad \text{to} \quad S \in \Omega \tag{17}$$

where $\Omega$ is the set of possible feature subsets, S refers to feature subsets.

The aim of three-class ROCCH maximization is to maximize the three objectives simultaneously. As the three objectives are in conflict with each other, the new proposed algorithm try to find the best trade-offs among them. It is possible to construct a classifier with two different classifiers with subsets of features in 3-D ROC space by means of randomization with the theory of iso-performance [1], which has computation complexity between the two classifiers.

TABLE IX
VUS. MEAN AND STANDARD DEVIATION

| | NSGA-II | SPEA2 | SMS-EMOA | 3DCH-EMOA |
|---|---|---|---|---|
| australian | $3.34e-01_{1.7e-02}$ | $4.01e-01_{1.4e-02}$ | $3.80e-01_{1.4e-02}$ | $4.25e-01_{8.8e-03}$ |
| breast-w | $3.79e-01_{1.8e-02}$ | $4.24e-01_{1.5e-02}$ | | $4.66e-01_{2.3e-03}$ |
| credit-a | $3.25e-01_{1.6e-02}$ | $3.61e-01_{1.8e-02}$ | $3.49e-01_{1.1e-02}$ | $4.10e-01_{8.4e-03}$ |
| cylinder-bands | $2.11e-01_{1.8e-02}$ | $3.38e-01_{1.2e-02}$ | $3.37e-01_{1.4e-02}$ | $3.53e-01_{2.1e-02}$ |
| diabetes | $2.11e-01_{2.0e-02}$ | $3.10e-01_{2.1e-02}$ | $3.01e-01_{2.7e-02}$ | $3.18e-01_{1.4e-02}$ |
| german | $1.81e-01_{1.8e-02}$ | $3.03e-01_{1.4e-02}$ | $2.95e-01_{1.1e-02}$ | $3.25e-01_{1.4e-02}$ |
| heart | $3.08e-01_{2.0e-02}$ | $3.90e-01_{1.4e-02}$ | $3.72e-01_{1.5e-02}$ | $4.21e-01_{5.7e-03}$ |
| hepatitis | $2.59e-01_{3.9e-02}$ | $3.20e-01_{1.7e-02}$ | $3.24e-01_{1.7e-02}$ | $3.83e-01_{1.4e-02}$ |
| hypothyroid_bi | $3.62e-01_{1.3e-02}$ | $3.63e-01_{1.5e-02}$ | $3.65e-01_{1.6e-02}$ | $4.26e-01_{1.5e-02}$ |
| labor | $3.08e-01_{3.3e-02}$ | $3.41e-01_{1.6e-02}$ | $3.40e-01_{1.6e-02}$ | $4.52e-01_{1.3e-02}$ |
| mushroom | $3.48e-01_{7.6e-03}$ | $3.47e-01_{8.1e-03}$ | $3.47e-01_{7.6e-03}$ | $4.66e-01_{1.3e-02}$ |
| ionosphere | $3.39e-01_{1.5e-02}$ | $3.74e-01_{1.4e-02}$ | $3.70e-01_{1.2e-02}$ | $4.63e-01_{1.3e-02}$ |
| kr-vs-kp | $3.61e-01_{1.1e-02}$ | $3.73e-01_{1.1e-02}$ | $3.61e-01_{1.1e-02}$ | $3.97e-01_{1.3e-02}$ |
| parkinsons | $3.22e-01_{2.8e-02}$ | $3.69e-01_{1.5e-02}$ | $3.62e-01_{1.2e-02}$ | $4.72e-01_{9.7e-03}$ |
| pima | $2.12e-01_{2.1e-02}$ | $3.10e-01_{1.9e-02}$ | $3.02e-01_{2.7e-02}$ | $3.18e-01_{1.4e-02}$ |
| sick | $3.56e-01_{1.7e-02}$ | $3.89e-01_{1.4e-02}$ | $3.82e-01_{1.5e-02}$ | $4.13e-01_{2.0e-02}$ |
| sonar | $2.94e-01_{1.3e-02}$ | $3.54e-01_{1.1e-02}$ | $3.52e-01_{9.8e-03}$ | $4.37e-01_{1.1e-02}$ |
| spambase | $3.53e-01_{1.4e-02}$ | $3.81e-01_{1.0e-02}$ | $3.72e-01_{1.1e-02}$ | $4.09e-01_{1.1e-02}$ |
| spect | $2.29e-01_{2.5e-02}$ | $2.68e-01_{2.3e-02}$ | $2.67e-01_{1.9e-02}$ | $2.97e-01_{2.5e-02}$ |
| spectf | $3.32e-01_{1.5e-02}$ | $3.67e-01_{1.1e-02}$ | $3.65e-01_{1.3e-02}$ | $4.49e-01_{1.1e-02}$ |
| tic-tac-toe | $2.66e-01_{1.2e-02}$ | $3.27e-01_{8.5e-03}$ | $3.24e-01_{1.0e-02}$ | $3.24e-01_{1.0e-02}$ |
| vote | $3.53e-01_{1.7e-02}$ | $3.65e-01_{1.3e-02}$ | $3.65e-01_{1.5e-02}$ | $4.55e-01_{6.6e-03}$ |
| wdbc | $3.56e-01_{1.6e-02}$ | $3.71e-01_{1.2e-02}$ | $3.74e-01_{1.5e-02}$ | $4.70e-01_{1.1e-02}$ |
| wpbc | $1.94e-01_{4.1e-02}$ | $3.42e-01_{2.7e-02}$ | $3.41e-01_{2.7e-02}$ | $3.87e-01_{3.2e-02}$ |

*1) Data sets:* Five data sets are selected from UCI repository [18] and described in Table X.

TABLE X
THREE-CLASS UCI DATA SETS

| No. | Data Set | No.of features | Class Distribution |
|---|---|---|---|
| 1 | cmc | 9 | 629:333:511 |
| 2 | connect-4 | 42 | 44473:16635:6449 |
| 3 | lung-cancer | 56 | 9:13:10 |
| 4 | waveform | 40 | 1692:1653:1655 |
| 5 | wine | 13 | 59:71:48 |

*2) Tested algorithms:* Similarly to the previous section, NSGA-II, MOEA/D, SPEA2 are involved to make experimental comparisons, and C4.5 [52] is used with J48 Java open source implementation of weka data mining tool [53].

*3) Configuration and Evaluation:* The goal of feature selection for three-class classification is to obtain large diverse subsets of features, which will lead to good performance of the classifiers ensemble. We use the VUS to measure the diversity of subsets of features in the 3-D ROC space.

The encoding of the chromosome and configuration of the algorithms are similarly to the presentation used in the previous Section VII-A.

*4) Results and Analysis:* The performance comparison of NSGA-II, SPEA2, SMS-EMOA and 3DCH-EMOA on feature selection for three-class classification with UCI data sets is discussed in this part. The results include the plots of 3-D ROC space and statistical analysis on VUS metric. The Pareto front of each algorithm of waveform data set are shown in Fig. 26, and the statistical results of the VUS are shown in Table.XI.

By comparing all the results we see the solutions of NSGA-II, SPEA2 and SMS-EMOA are centered on the surroundings of point (0,0,0), which means these algorithms do not have good performance on diversity. The new proposed method can obtain a solution space with higher value of VUS than others

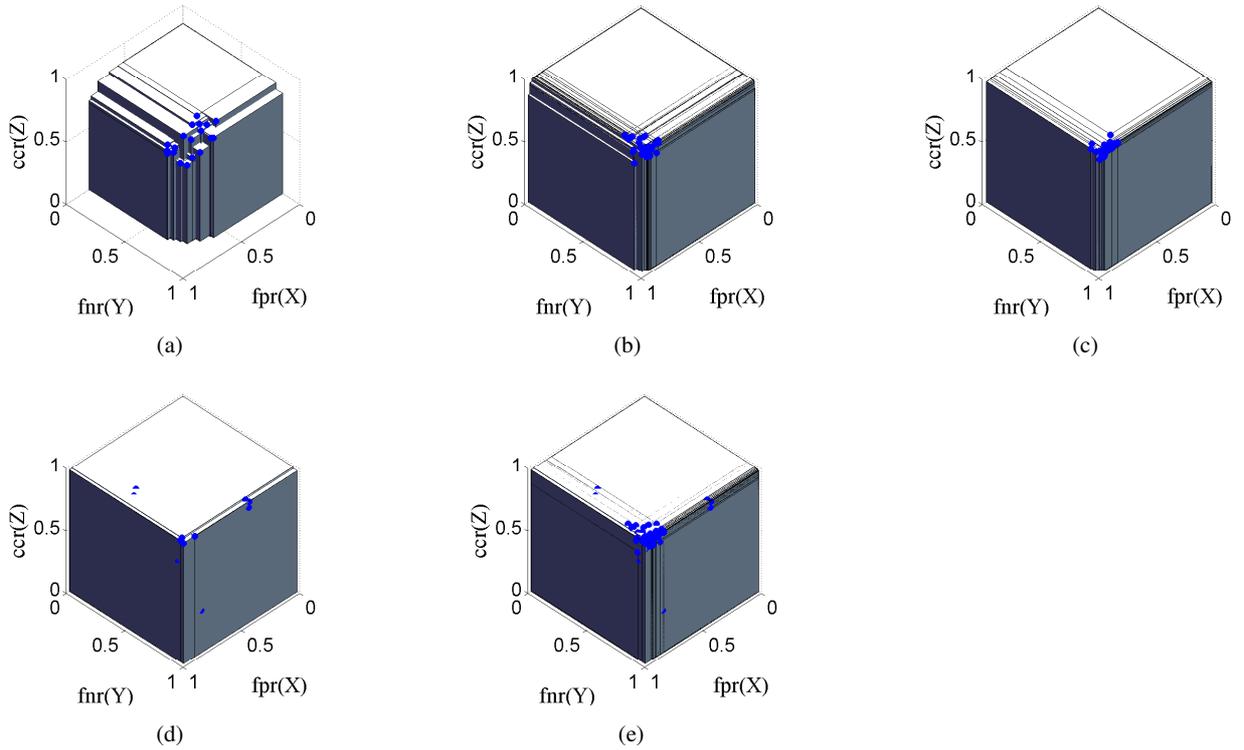

Fig. 26. Experimental results of feature selection of waveform data set are shown in $tar - tbr - tcr$ space. (a) Result of NSGA-II in 3D ROC space. (b) Result of SPEA2. (c) Result of SMS-EMOA. (d) Result of 3DCH-EMOA. (e) Reference Pareto Front.

<div align="center">

TABLE XI
VUS. MEAN AND STANDARD DEVIATION

</div>

|  | NSGA-II | SPEA2 | SMS-EMOA | 3DCH-EMOA |
|---|---|---|---|---|
| cmc | $1.13e-01_{2.7e-02}$ | $1.59e-01_{1.6e-02}$ | $1.60e-01_{1.6e-02}$ | $1.87e-01_{9.9e-03}$ |
| connect-4 | $2.19e-01_{1.0e-02}$ | $2.59e-01_{9.4e-03}$ | $2.57e-01_{1.1e-02}$ | $2.57e-01_{1.7e-02}$ |
| lung-cancer | $3.01e-01_{7.7e-02}$ | $3.51e-01_{1.4e-02}$ | $3.56e-01_{1.1e-02}$ | $4.04e-01_{2.1e-02}$ |
| waveform-5000 | $2.01e-01_{7.5e-03}$ | $3.27e-01_{6.6e-03}$ | $3.26e-01_{9.9e-03}$ | $3.50e-01_{2.5e-02}$ |
| wine | $3.36e-01_{1.7e-02}$ | $3.33e-01_{6.0e-04}$ | $3.33e-01_{7.4e-04}$ | $3.72e-01_{2.3e-02}$ |

for most of the results, which is shown in Table. XI. From the Fig. 26(d) we can see that the solutions of 3DCH-EMOA are more scattered than other algorithms, which results in good performance of ensemble learning.

## VIII. CONCLUSIONS AND FUTURE WORK

In this paper, we analyzed the properties of augmented and 3-D ROC. 3DCH-EMOA is proposed to optimize the performance of 3-D ROCCH for classification. In order to evaluate the performance of several evolutionary multiobjective algorithms two set of test problems, ZEJD and ZED were designed. 3DCH-EMOA is compared with other EMOAs, such as NSGA-II, GDE3, SPEA2, MOEA/D and SMS-EMOA on ZEJD and ZED test problems. 3DCH-EMOA always obtain the best results not only in convergence but also in diversity. 3DCH-EMOA revealed a good ability to heal the dent in 3-D ROC space, which is really important in ROCCH maximization problems. We also applied this algorithm to the area of SPAM filtering optimization and feature selection. Extensive experimental studies compared the new proposed method with state-of-the-art approaches, proving that the proposed algorithm is promising and effective.

However, the new proposed method is time consuming, because it needs to compute the VUS contribution of every point in the first priority level solutions. In the future, the dynamic convex hulls strategy will be adapted to reduce the computational complexity.

## Acknowledgment

This work was partially supported by the National Basic Research Program (973 Program) of China (No. 2013CB329402), the National Natural Science Foundation of China (No. 61273317, 61271301, 61272279, 61001202, 61203303, 61003199 and 61003198), The Fund for Foreign Scholars in University Research and Teaching Programs (the 111 Project) (No. B07048), the National Research Foundation for the Doctoral Program of Higher Education of China (No. 20100203120008), the Program for Cheung Kong Scholars and Innovative Research Team in University (No. IRT1170), the Natural Science Basic Research Plan in Shaanxi Province of China (Program No.2014JM8301), and the European Union Seventh Framework Programme under Grant agreements (No. 247619) on Nature Inspired Computation and its Applications (NICaiA).